\pdfoutput=1

\documentclass[sigconf]{acmart}

\usepackage{caption}
\usepackage{setspace}
\usepackage{enumitem}

\usepackage{subcaption}
\usepackage{hyperref}
\usepackage{bbm}
\usepackage{algorithm}
\usepackage{algorithmic}
\usepackage{tabularx}

\usepackage{diagbox}
\usepackage{flushend}

\def \eg {\emph{e.g.}, }
\def \ie {\emph{i.e.}, }
\def \th {\mathop{\mathrm{th}}}

\begin{document}

\title[Compressing Deep Neural Network Structures for Sensing Systems]{DeepIoT: Compressing Deep Neural Network Structures for Sensing Systems with a Compressor-Critic Framework}


\copyrightyear{2017} 
\acmYear{2017} 
\setcopyright{acmcopyright}
\acmConference{SenSys '17}{November 6--8, 2017}{Delft, Netherlands}\acmPrice{15.00}\acmDOI{10.1145/3131672.3131675}
\acmISBN{978-1-4503-5459-2/17/11}


\author{Shuochao Yao}
\affiliation{%
  \institution{University of Illinois Urbana Champaign}
  }
  
\author{Yiran Zhao}
\affiliation{%
  \institution{University of Illinois Urbana Champaign}
  } 
 
\author{Aston Zhang}
\affiliation{%
 \institution{University of Illinois Urbana Champaign}
  }
  
\author{Lu Su}
\affiliation{%
  \institution{State University of New York at Buffalo}
  }

\author{Tarek Abdelzaher}
\affiliation{%
  \institution{University of Illinois Urbana Champaign}
  }

\sloppy

\begin{abstract}
Recent advances in deep learning motivate the use of deep neutral networks in sensing applications, but their excessive resource needs on constrained embedded devices remain an important impediment. A recently explored solution space lies in compressing (approximating or simplifying) deep neural networks in some manner before use on the device.
We propose a new compression solution, called DeepIoT, that makes two key contributions in that space. First, unlike current solutions geared for compressing specific types of neural networks, DeepIoT presents
a unified approach that compresses all commonly used deep learning structures for sensing applications, including fully-connected, convolutional, and recurrent neural networks, as well as their combinations. Second, unlike solutions that either sparsify weight matrices or assume linear structure within weight matrices,
DeepIoT compresses neural network structures into smaller dense matrices by finding the minimum number of non-redundant hidden elements, such as filters and dimensions required by each layer, while keeping the performance of sensing applications the same. Importantly, it does so using an approach that obtains a global view of parameter redundancies, which is shown to produce superior compression.
The compressed model generated by DeepIoT can directly use existing deep learning libraries that run on embedded and mobile systems without further modifications. 
We conduct experiments with five different sensing-related tasks on Intel Edison devices. DeepIoT outperforms all compared baseline algorithms with respect to execution time and energy consumption by a significant margin. It reduces the size of deep neural networks by $90\%$ to $98.9\%$. It is thus able to shorten execution time by $71.4\%$ to $94.5\%$, and decrease energy consumption by $72.2\%$ to $95.7\%$. These improvements are achieved without loss of accuracy. The results underscore the potential of DeepIoT for advancing the exploitation of deep neural networks on resource-constrained embedded devices.
\end{abstract}

\maketitle

{

\vspace{-0.2cm}
\section{Introduction}~\label{sec:intro}
This paper is motivated by the prospect of enabling a ``smarter" and more user-friendly category of every-day physical objects capable of performing complex sensing and recognition tasks, such as those needed for understanding human context and enabling more natural interactions with users in emerging Internet of Things (IoT) applications.

Present-day sensing applications cover a broad range of areas 
including human interactions~\cite{zhang2009ocrdroid,hoque2014vocal}, context sensing~\cite{capra2003carisma,yang2011detecting,nirjon2012musicalheart,rowe2010contactless,sun2011pandaa}, crowd sensing~\cite{yao2016recursive,zhang2017regions}, object detection and tracking~\cite{shen2014face,wilson2011see,cho2011inferring,kusy2007tracking}. 
The recent commercial interest in IoT technologies promises a proliferation of smart objects in human spaces at a much broader scale. Such objects will ideally have independent means of interacting with their surroundings to perform complex detection and recognition tasks, such as recognizing users, interpretting voice commands, and understanding human context. The paper explores the feasibility of implementing such functions using deep neural networks on computationally-constrained devices, such as Intel's suggested IoT platform: the Edison board\footnote{https://software.intel.com/en-us/iot/hardware/edison}.

The use of deep neural networks in sensing applications has recently gained popularity. Specific neural network models have been designed to fuse multiple sensory modalities and extract temporal relationships for sensing applications. These models have shown significant improvements on audio sensing~\cite{lane2015deepear}, tracking and localization~\cite{clark2017vinet,rosa2017leveraging,wang2017deepvo,yao2017deepsense}, human activity recognition~\cite{radu2016towards,yao2017deepsense}, and user identification~\cite{lane2015deepear,yao2017deepsense}.

Training the neural network can occur on a computationally capable node and, as such, is not of concern in this paper. The key impediment to deploying deep-learning-based sensing applications lies in the high
memory consumption, execution time, and energy demand associated with storing and using the {\em trained network\/} on the {\em target device\/}. 
This leads to increased interest in compressing neural networks 
to enable exploitation of deep learning on low-end embedded devices.  
                         
We propose DeepIoT that compresses commonly used deep neural network structures for sensing applications 
through deciding the minimum number of elements in each layer. 
Previous illuminating studies on neural network compression sparsify large dense parameter matrices into large sparse matrices~\cite{han2015deep,guo2016dynamic,lane2016sparsifying}. In contrast, DeepIoT minimizes the number of elements in each layer, which results in converting parameters into a set of small dense matrices. A small dense matrix does not require additional storage for element indices and is efficiently optimized for processing~\cite{goumas2008understanding}. DeepIoT greatly reduces the effort of designing efficient neural structures for sensing applications by deciding the number of elements in each layer in a manner informed by the topology of the neural network.
                           
DeepIoT borrows the idea of dropping hidden elements from a widely-used deep learning regularization method called dropout~\cite{srivastava2014dropout}. 
The dropout operation gives each hidden element a dropout probability. 
During the dropout process, hidden elements can be pruned according to their dropout probabilities. Then a ``thinned" network structure can be generated. However, these dropout probabilities are usually set to a pre-defined value, such as 0.5. Such pre-defined values are not the optimal probabilities, thereby resulting in a less efficient exploration of the solution space. If we can obtain the optimal dropout probability for each hidden element, it becomes possible for us to generate the optimal slim network structure that preserves the accuracy of sensing applications while maximally reducing the resource consumption of sensing systems. An important purpose of DeepIoT is thus to find the optimal dropout probability for each hidden element in the neural network. 

Notice that, dropout can be easily applied to all commonly used neural network structures. In fully-connected neural networks, neurons are dropped in each layer~\cite{srivastava2014dropout}; in convolutional neural networks, filters are dropped in each layer~\cite{gal2015bayesian}; and in recurrent neural networks, dimensions are reduced in each layer~\cite{gal2015theoretically}. This means that DeepIoT can be applied to all commonly-used neural network structures and their combinations.

To obtain the optimal dropout probabilities for the neural network, DeepIoT exploits the network parameters themselves. From the perspective of model compression, a hidden element that is connected to redundant model parameters should have a higher probability to be dropped. 
A contribution of DeepIoT lies in exploiting a novel {\em compressor\/} neural network to solve this problem. 
It takes model parameters of each layer as input, learns parameter redundancies, and generates the dropout probabilities accordingly. Since there are interconnections of parameters among different layers, we design the compressor neural network to be a recurrent neural network that can globally share the redundancy information and generate dropout probabilities layer by layer. 

The compressor neural network is optimized 
jointly with the original neural network to be compressed through a compressor-critic framework that tries to minimize the loss function of the original sensing applicaiton. The compressor-critic framework emulates the idea of the well-known actor-critic algorithm from reinforcement learning~\cite{konda1999actor}, optimizing two networks in an iterative manner. 

We evaluate the DeepIoT framework on the Intel Edison computing platform~\cite{Edison}, which Intel markets as an enabler platform for the computing elements of embedded ``things" in IoT systems. We conduct two sets of experiments. 
The first set consists of three tasks that enable embedded systems to interact with humans with basic modalities, including handwritten text, vision, and speech, demonstrating superior accuracy of our produced neural networks, compared to others of similar size.
The second set provides two examples of applying compressed neural networks to solving human-centric context sensing tasks; namely, human activity recognition and user identification, in a resource-constrained scenario. 

We compare DeepIoT with other state-of-the-art magnitude-based~\cite{guo2016dynamic} and sparse-coding-based~\cite{lane2016sparsifying} neural network compression methods. The resource consumption of resulting models on the Intel Edison module and the final performance of sensing applications are estimated for all compressed models. In all experiments, DeepIoT is shown to outperform the other algorithms by a large margin in terms of compression ratio, memory consumption, execution time, and energy consumption. In these experiments, when compared with the non-compressed neural networks, DeepIoT is able to reduce the model size by 
$90\%$ to $98.9\%$, shorten the running time by 
$71.4\%$ to $94.5\%$, and decrease the energy consumption by 
$72.2\%$ to $95.7\%$. 
When compared with the state-out-the-art baseline algorithm, DeepIoT is able to reduce the model size by 
$11.6\%$ to $83.2\%$, shorten the running time by 
$60.9\%$ to $87.9\%$, and decrease the energy consumption by 
$64.1\%$ to $88.7\%$. 
Importantly, these improvements are achieved without loss of accuracy. Experiments demonstrate the promise of DeepIoT in enabling resource-constrained embedded devices to benefit from advances in deep learning.

The rest of this paper is organized as follows. Section~\ref{sec:related} introduces related work on optimizating sensing applications for resource-constrained devices. We describe the technical details of DeepIoT in Section~\ref{sec:model}. We describe system implementation in Section~\ref{sec:implementation}. The evaluation is presented in Section~\ref{sec:evaluation}. Finally, we discuss the results in Section~\ref{sec:discussion} and conclude in Section~\ref{sec:conclusion}.

\vspace{-0.1cm}
\section{Related Work}~\label{sec:related}

A key direction in embedded sensing literaure is to enable running progressively more interesting applications under the  more pronounced resource constraints of embedded and mobile devices. Brouwers et al. reduced the energy consumption of Wi-Fi based localization with an incremental scanning strategy~\cite{brouwers2014incremental}. Hester et al. proposed an ultra-low-power hardware architecture and a companion software framework for energy-efficient sensing system~\cite{hester2016amulet}. Ferrari et al. and Schuss et al. focused on low-power wireless communication protocols~\cite{ferrari2012low,schuss2017competition}. Wang et al. enabled energy efficient reliable broadcast by considering poorly correlated links~\cite{wang2013corlayer}. Saifullah et al. designed a scalable and energy-efficient wireless sensor network (WSN) over white spaces~\cite{saifullah2016snow}. Alsheikh et al. discussed about data compression in WSN with machine learning techniques~\cite{alsheikh2014machine}.

\begin{figure}[!htb]
\centering
\includegraphics[width=0.75\linewidth]{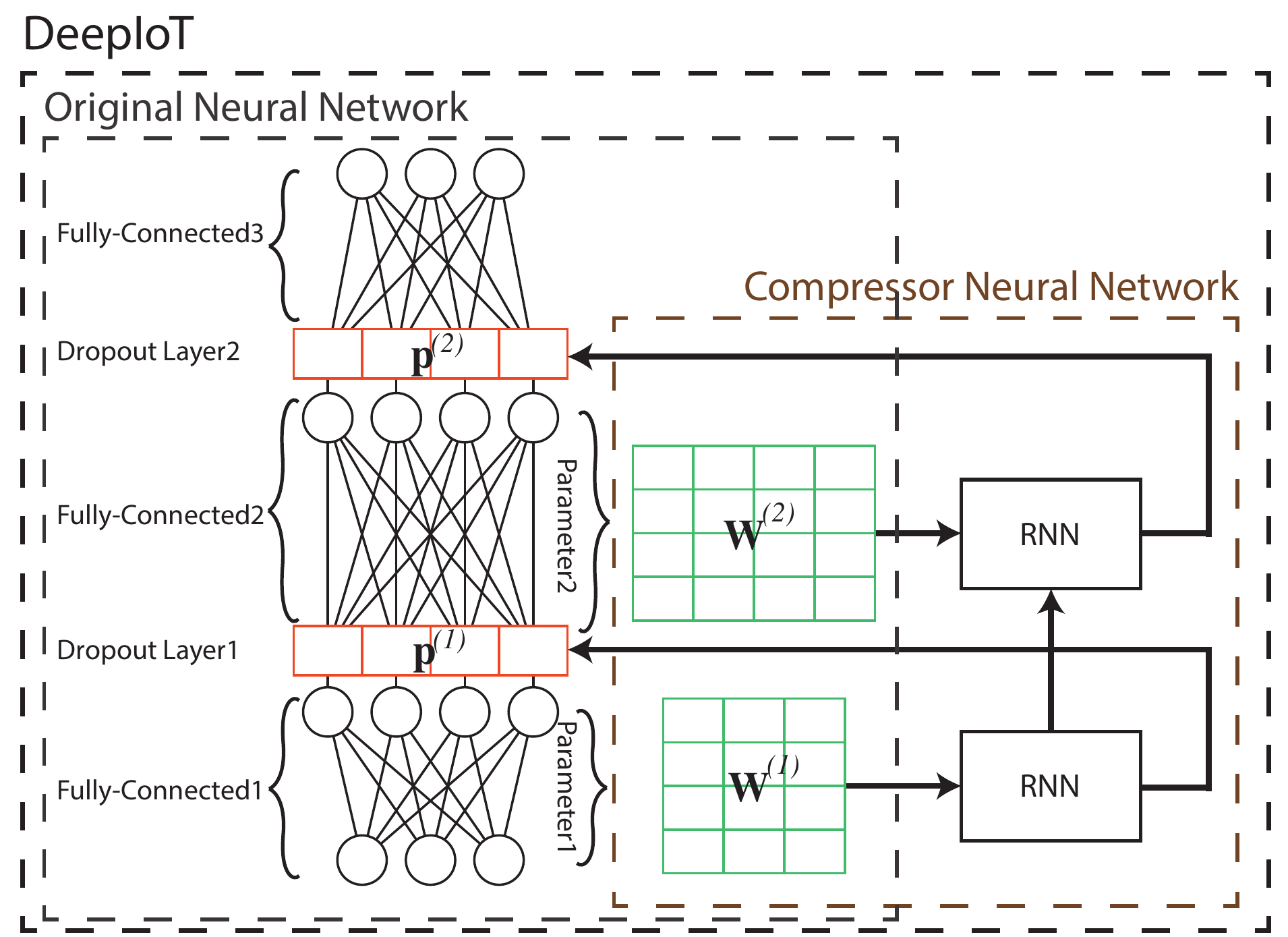}
\vspace{-0.1cm}
\caption{Overall DeepIoT system framework. Orange boxes represent dropout operations. Green boxes represent parameters of the original neural network.}
\label{fig:overall}
\vspace{-0.3cm}
\end{figure}

Recent studies focused on compressing deep neural networks for embedded and mobile devices.
Han et al. proposed a magnitude-based compression method with fine-tuning, which illustrated promising compression results~\cite{han2015deep}. This method removes weight connections with low magnitude iteratively; however, it requires additional implementation of sparse matrix with more resource consumption. In addition, the aggressive pruning method increases the potential risk of irretrievable network damage. Guo et al. proposed a compression algorithm with connection splicing, which provided the chance of rehabilitation with a certain threshold~\cite{guo2016dynamic}. However, the algorithm still focuses on weight level instead of structure level. 
Other than the magnitude-based method,  another series of works focused on the factorization-based method that reduced the neural network complexity by exploiting low-rank structures in parameters. Denton et al. exploited various matrix factorization methods with fine-tunning to approximate the convolutional operations in order to reduce the neural network execution time~\cite{denton2014exploiting}. Lane et al. applied sparse coding based and matrix factorization based method to reduce complexity of fully-connected layer and convolutional layer respectively~\cite{lane2016sparsifying}. However, factorization-based methods usually obtain lower compression ratio compared with magnitude-based methods, and the low-rank assumption may hurt the final network performance. 
Wang et al. applied the information of frequency domain for model compression~\cite{wang2016cnnpack}. However, additional implementation is required to speed-up the frequency-domain representations, and the method is not suitable for modern CNNs with small convolution filter sizes.
 Hinton et al. proposed a teacher-student framework that distilled the knowledge in an ensemble of models into a single model~\cite{hinton2015distilling}. However, the framework focused more on compressing model ensemble into a single model instead of structure compression.

Our paper is partly inspired by deep reinforcement learning. With the aid of deep neural networks, reinforcement leaning has achieved great success on Atari games~\cite{mnih2013playing}, Go chess~\cite{silver2016mastering}, 
and multichannel access~\cite{wang2017deep}. 

To the best of our knowledge, DeepIoT is the first framework for neural network structure compressing based on dropout operations and reducing parameter redundancies, where dropout operations provide DeepIoT the chance of rehabilitation with a certain probability. DeepIoT generates a more concise network structure for transplanting large-scale neural networks onto resource-constrained embedded devices.

\section{System Framework}~\label{sec:model}
We introduce DeepIoT, a neural network structure compression framework for sensing applications.
Without loss of generality, before introducing the technical details, we first use an example of compressing a 3-layer fully-connected neural network structure to illustrate the overall pipeline of DeepIoT.
The detailed illustration is shown in Figure~\ref{fig:overall}. The basic steps of compressing neural network structures for sensing applications with DeepIoT can be summarized as follows.
\begin{enumerate}[leftmargin=*]
\setlength\itemsep{0pt}
\item 
Insert operations that randomly zeroing out hidden elements with probabilities $\mathbf{p}^{(l)}$ called dropout
(red boxes in Figure~\ref{fig:overall}) into internal layers of the original neural network. The internal layers exclude input layers and output layers that have the fixed dimension for a sensing application. This step will be detailed in Section~\ref{sec:dropout}.
\item Construct the compressor neural network. It takes the weight matrices $\mathbf{W}^{(l)}$ (green boxes in Figure~\ref{fig:overall}) from the layers to be compressed in the original neural network as inputs, learns and shares the parameter redundancies among different layers,  and generates optimal dropout probabilities $\mathbf{p}^{(l)}$, which is then fed back to the dropout operations in the original neural network. This step will be detailed in Section~\ref{sec:compressor}.
\item Iteratively optimize the compressor neural network and the original neural network with the compressor-critic framework. The compressor neural network is optimized to produce better dropout probabilities that can generate a more efficient network structure for the original neural network. The original neural network is optimized to achieve a better performance with the more efficient  structure for a sensing application. This step will be detailed in Section~\ref{sec:compressor_critic}.
\end{enumerate}

For the rest of this paper, all vectors are denoted by bold lower-case letters (\eg $\mathbf{x}$ and $\mathbf{y}$), and matrices and tensors are represented by bold upper-case letters (\eg $\mathbf{X}$ and $\mathbf{Y}$). For a column vector $\mathbf{x}$, the $j^{\th}$ element is denoted by $\mathbf{x}_{[j]}$. For a tensor $\mathbf{X}$, the $t^{\th}$ matrix along the third axis is denoted by $\mathbf{X}_{\cdot \cdot t}$, and the other slicing denotations are defined similarly. The superscript $l$ in $\mathbf{x}^{(l)}$ and $\mathbf{X}^{(l)}$ denote the vector and tensor for the $l^{\th}$ layer of the neural network. We use calligraphic letters to denote sets (\eg $\mathcal{X}$ and $\mathcal{Y}$). For any set $\mathcal{X}$, $|\mathcal{X}|$ denotes the cardinality of $\mathcal{X}$.

\subsection{Dropout Operations in the Original Neural Network}~\label{sec:dropout}
Dropout is commonly used as a regularization method that prevents feature co-adapting and model overfitting.  The term ``dropout" refers to dropping out units (hidden and visible) in a neural network. Since DeepIoT is a structure compression framework, we focus mainly on dropping out hidden units. The definitions of hidden units are distinct in different types of neural networks, and we will describe them in detail. The basic idea is that we regard neural networks with dropout operations as bayesian neural networks with Bernoulli variational distributions~\cite{gal2015bayesian, gal2015theoretically, srivastava2014dropout}.

For the fully-connected neural networks, the fully-connected operation with dropout can be formulated as
%
\begin{equation}
\small
\begin{split}
\mathbf{z}_{[j]}^{(l)} &\sim  \text{Bernoulli}(\mathbf{p}_{[j]}^{(l)}), \\
\tilde{\mathbf{W}}^{(l)} &=   \mathbf{W}^{(l)}\text{diag}\big( \mathbf{z}^{(l)}\big), \\
\mathbf{Y}^{(l)} &=\mathbf{X}^{(l)}  \tilde{\mathbf{W}}^{(l)}  + \mathbf{b}^{(l)},\\
\mathbf{X}^{(l+1)} &= f\big(\mathbf{Y}^{(l)}\big).
\end{split}
\label{eqn:dropout}
\end{equation}
%

Refer to \eqref{eqn:dropout}. The notation $l = 1, \cdots, L$ is the layer number in the fully-connected neural network. For any layer $l$, the weight matrix is denoted as $\mathbf{W}^{(l)}\in \mathbb{R}^{d^{(l-1)}\times d^{(l)}}$; the bias vector is denoted as $\mathbf{b}^{(l)}\in \mathbb{R}^{d^{(l)}}$; and the input is denoted as $\mathbf{X}^{(l)} \in \mathbb{R}^{1 \times d^{(l-1)}}$. In addition, $f(\cdot)$ is a nonlinear activation function. 

As shown in \eqref{eqn:dropout}, each hidden unit is controlled by a Bernoulli random variable. In the original dropout method, the success probabilities of $\mathbf{p}_{[j]}^{(l)}$ can be set to the same constant $p$ for all hidden units~\cite{srivastava2014dropout}, but DeepIoT uses the Bernoulli random variable with individual success probabilities for different hidden units in order to compress the neural network structure in a finer granularity.

For the convolutional neural networks, the basic fully-connected operation is replaced by the convolution operation~\cite{gal2015bayesian}. However, the convolution can be reformulated as a linear operation as shown in  \eqref{eqn:dropout}. For any layer $l$, we denote  $\mathcal{K}^{(l)}=\big\{\mathbf{K}_k^{(l)}\big\}$ for $k = 1, \cdots, c^{(l)}$ as the set of convolutional neural network (CNN)'s kernels, where $\mathbf{K}_k^{(l)} \in \mathbb{R}^{h^{(l)}\times w^{(l)}\times c^{(l-1)}}$ is the kernel of CNN with height $h^{(l)}$, width $w^{(l)}$, and channel $c^{(l-1)}$. The input tensor of layer $l$ is denoted as $\mathbf{\hat{X}}^{(l)} \in \mathbb{R}^{\hat{h}^{(l-1)}\times \hat{w}^{(l-1)}\times c^{(l-1)}}$ with height $\hat{h}^{(l-1)}$, width $\hat{w}^{(l-1)}$, and channel $c^{(l-1)}$. 

Next, we convert convolving the kernels with the input into performing matrix product. We extract $h^{(l)}\times w^{(l)}\times c^{(l-1)}$ dimensional patches from the input $\mathbf{\hat{X}}^{(l)}$ with stride $s$ and vectorize them. Collect these vectorized $n$ patches to be the rows of our new input representation $\mathbf{X}^{(l)}\in \mathbb{R}^{n\times (h^{(l)}w^{(l)}c^{(l-1)})}$. The vectorized kernels form the columns of the weight matrix $\mathbf{W}^{(l)} \in \mathbb{R}^{(h^{(l)}w^{(l)}c^{(l-1)}) \times c^{(l)}}$. 

With this transformation, dropout operations can be applied to convolutional neural networks according to \eqref{eqn:dropout}. The composition of pooling and activation functions can be regarded as the nonlinear function $f(\cdot)$ in \eqref{eqn:dropout}. Instead of dropping out hidden elements in each layer, we drop out convolutional kernels in each layer. From the perspective of structure compression, DeepIoT  tries to prune the number of kernels used in the convolutional neural networks.

For the recurrent neural network, we take a multi-layer Long Short Term Memory network (LSTM) as an example. The LSTM operation with dropout can be formulated as
\begin{equation}
\small
\begin{split}
\mathbf{z}_{[j]}^{(l)} &\sim  \text{Bernoulli}(\mathbf{p}_{[j]}^{(l)}), \\
\begin{pmatrix} \mathbf{i} \\\mathbf{f} \\ \mathbf{o} \\ \mathbf{g} \end{pmatrix}  & =
\begin{pmatrix} \text{sigm} \\ \text{sigm} \\ \text{sigm} \\ \text{tanh} \end{pmatrix}
\mathbf{W}^{(l)}
\begin{pmatrix} \mathbf{h}_t^{(l-1)} \odot \mathbf{z}^{(l-1)} \\ \mathbf{h}_{t-1}^{(l)} \odot \mathbf{z}^{(l)} \end{pmatrix},\\
\mathbf{c}_{t}^{(l)} &= \mathbf{f} \odot \mathbf{c}_{t-1}^{(l)} + \mathbf{i} \odot \mathbf{g},\\
\mathbf{h}_{t}^{(l)} &= \mathbf{o} \odot \text{tanh}\big( \mathbf{c}_{t}^{(l)}  \big).
\end{split}
\label{eqn:LSTM_Dropout}
\end{equation}

The notation $l = 1,\cdots ,L$ is the layer number and $t = 1,\cdots, T$ is the step number in the recurrent neural network. Element-wise multiplication is denoted by $\odot$. Operators $sigm$ and $tanh$ denote sigmoid function and hyperbolic tangent respectively. The vector $\mathbf{h}_{t}^{(l)} \in\mathbb{R}^{n^{(l)}}$ is the output of step $t$ at layer $l$. The vector $\mathbf{h}_t^{(0)} = \mathbf{x}_t$ is the input for the whole neural network at step $t$. The matrix $\mathbf{W}^{(l)} \in \mathbb{R}^{4n^{(l)}\times (n^{(l-1)} + n^{(l)})}$ is the weight matrix at layer $l$. We let $\mathbf{p}_{[j]}^{(0)} = 1$, since DeepIoT only drops hidden elements.

As shown in \eqref{eqn:LSTM_Dropout}, DeepIoT uses the same vector of Bernoulli random variables $\mathbf{z}^{(l)}$ to control the dropping operations among different time steps in each layer, while individual Bernoulli random variables are used for different steps in the original LSTM dropout~\cite{zaremba2014recurrent}. From the perspective of structure compression, DeepIoT  tries to prune the number of hidden dimensions used in LSTM blocks. The dropout operation of other recurrent neural network architectures, such as Gated Recurrent Unit (GRU), can be designed similarly.  

\subsection{Compressor Neural Network}~\label{sec:compressor}
Now we introduce the architecture of the compressor neural network. As we described in Section~\ref{sec:intro}, a hidden element in the original neural network that is connected to redundant model parameters should have a higher probability to be dropped. Therefore we design the compressor neural network to take the weights of an original neural network $\{\mathbf{W}^{(l)}\}$ as inputs, learn the redundancies among these weights, and generate  
dropout probabilities $\{\mathbf{p}^{(l)}\}$ for
hidden elements that can be eventually used to compress the original neural network structure.

A straightforward solution is to train an individual fully-connected neural network for each layer in the original neural network. However, since there are interconnections among weight redundancies in different layers, DeepIoT uses a variant LSTM as the structure of compressor to share and use the parameter redundancy information among different layers.

According to the description in Section~\ref{sec:dropout}, the weight in layer $l$ of fully-connected, convolutional, or recurrent neural network can all be represented as a single matrix $\mathbf{W}^{(l)}\in \mathbb{R}^{d_{f}^{(l)}\times d_{\rm{drop}}^{(l)}}$, where $d_{\rm{drop}}^{(l)}$ denotes the dimension that dropout operation is applied and $d_{f}^{(l)}$ denotes the dimension of features within each dropout element. 
Here, we need to notice that the weight matrix of LSTM at layer $l$ can be reshaped as $\mathbf{W}^{(l)} \in \mathbb{R}^{4\cdot(n^{(l-1)} + n^{(l)})\times n^{(l)}}$, where $d_{\rm{drop}}^{(l)} = n^{(l)}$ and $d_{f}^{(l)} = 4\cdot(n^{(l-1)} + n^{(l)})$.
Hence, we take weights from the original network layer by layer, $\mathcal{W} = \big\{\mathbf{W}^{(l)}\big\}$ with $l = 1, \cdots, L$, as the input of the compressor neural network.
Instead of using a vanilla LSTM as the structure of compressor, we apply a variant $l$-step LSTM model shown as 
\begin{equation}
\small
\begin{split}
\begin{pmatrix} \mathbf{v}_i^\intercal \\\mathbf{v}_f^\intercal \\ \mathbf{v}_o^\intercal \\ \mathbf{v}_g^\intercal \end{pmatrix}  & = \mathbf{W}_c^{(l)} \mathbf{W}^{(l)} \mathbf{W}_i^{(l)},  \quad
\begin{pmatrix} \mathbf{u}_i \\\mathbf{u}_f\\ \mathbf{u}_o \\ \mathbf{u}_g\end{pmatrix}   = \mathbf{W}_h \mathbf{h}_{l-1},  \\
\begin{pmatrix} \mathbf{i} \\\mathbf{f} \\ \mathbf{o} \\ \mathbf{g} \end{pmatrix}  & =
\begin{pmatrix} \text{sigm} \\ \text{sigm} \\ \text{sigm} \\ \text{tanh} \end{pmatrix}
\left( \begin{pmatrix} \mathbf{v}_i \\\mathbf{v}_f\\ \mathbf{v}_o \\ \mathbf{v}_g\end{pmatrix} +
\begin{pmatrix} \mathbf{u}_i \\\mathbf{u}_f\\ \mathbf{u}_o \\ \mathbf{u}_g\end{pmatrix}  \right), \\
\mathbf{c}_{l} &= \mathbf{f} \odot \mathbf{c}_{l-1} + \mathbf{i} \odot \mathbf{g},\\
\mathbf{h}_{l} &= \mathbf{o} \odot \text{tanh}\big( \mathbf{c}_{l}  \big),\\
\mathbf{p}^{(l)} &= \mathbf{p}_{t}= \text{sigm}\big(\mathbf{W}_o^{(l)} \mathbf{h}_{l} \big),\\
\mathbf{z}_{[j]}^{(l)} &\sim \text{Bernoulli}(\mathbf{p}_{[j]}^{(l)}).
\end{split}
\label{eqn:compressor}
\end{equation}

Refer to \eqref{eqn:compressor}, we denote $d_c$ as the dimension of the variant LSTM hidden state. Then $\mathbf{W}^{(l)} \in \mathbb{R}^{d_{f}^{(l)}\times d_{\rm{drop}}^{(l)}}$, $\mathbf{W}_c^{(l)} \in \mathbb{R}^{4\times d_{f}^{(l)}}$, $\mathbf{W}_i^{(l)} \in \mathbb{R}^{d_{\rm{drop}}^{(l)}\times d_{c}}$, $\mathbf{W}_h \in \mathbb{R}^{4d_{c}\times d_{c}}$, and $\mathbf{W}_o^{(l)} \in \mathbb{R}^{d_{\rm{drop}}^{(l)}\times d_{c}}$. The set of training parameters of the compressor neural network is denoted as $\phi$, where $\phi = \big\{\mathbf{W}_c^{(l)}, \mathbf{W}_i^{(l)}, \mathbf{W}_h, \mathbf{W}_o^{(l)}\big\}$. The matrix $\mathbf{W}^{(l)}$ is the input matrix for step $l$ in the compressor neural network, which is also the $l^{\th}$ layer's parameters of the original neural network in \eqref{eqn:dropout} or \eqref{eqn:LSTM_Dropout}.

Compared with the vanilla LSTM that requires vectorizing the original weight matrix as inputs, the variant LSTM model preserves the structure of original weight matrix and uses less learning parameters to extract the redundancy information among the dropout elements. In addition, $\mathbf{W}_c^{(l)}$ and $\mathbf{W}_i^{(l)}$ convert original weight matrix $\mathbf{W}^{(l)}$ with different sizes into fixed-size representations. The binary vector $\mathbf{z}^{(l)}$ is the dropout mask and probability $\mathbf{p}^{(l)}$ is the dropout probabilities for the $l^{\th}$ layer in the original neural network used in \eqref{eqn:dropout} and \eqref{eqn:LSTM_Dropout}, which is also the stochastic dropout policy learnt through observing the weight redundancies of the original neural network. 

\subsection{Compressor-Critic Framework}~\label{sec:compressor_critic}
In Section~\ref{sec:dropout} and Section~\ref{sec:compressor}, we have introduced customized dropout operations applied on the original neural networks that need to be compressed and the structure of compressor neural network used to learn dropout probabilities based on parameter redundancies. In this subsection, we will discuss the detail of compressor-critic compressing process. It optimizes the original neural network and the compressor neural network in an iterative manner and enables the compressor neural network to gradually compress the original neural network with soft deletion.

We denote the original neural network as $F_{\mathcal{W}}(\mathbf{x}|\mathbf{z})$, and we call it critic. It takes $\mathbf{x}$ as inputs and generates predictions based on binary dropout masks $\mathbf{z}$ and model parameters $\mathcal{W}$ that refer to a set of weights $\mathcal{W} = \{\mathbf{W}^{(l)}\}$ . We assume that $F_{\mathcal{W}}(\mathbf{x}|\mathbf{z})$ is a pre-trained model. We denote the compressor neural network by $\mathbf{z} \sim \mu_{\phi}(\mathcal{W})$. It takes the weights of the critic as inputs and generates the probability distribution of the mask vector $\mathbf{z}$ based on its own parameters $\phi$. In order to optimize the compressor to drop out hidden elements in the critic, DeepIoT follows the objective function
\begin{equation}
\small
\begin{split}
\mathcal{L} &= \mathbb{E}_{\mathbf{z} \sim \mu_{\phi}} \big[ L\big( \mathbf{y}, F_{\mathcal{W}}(\mathbf{x}|\mathbf{z})\big)  \big] \\
& = \sum_{\mathbf{z} \sim \{0,1\}^{|\mathbf{z}|}} \mu_{\phi}(\mathbf{W}) \cdot L\big(\mathbf{y} , F_{\mathcal{W}}(\mathbf{x}|\mathbf{z})\big),
\end{split}
\label{eqn:org_objective}
\end{equation}
where $L(\cdot, \cdot)$ is the objective function of the critic. The objective function can be interpreted as the expected loss of the original neural network over the dropout probabilities generated by the compressor.

DeepIoT optimizes the compressor and critic in an iterative manner. It reduces the expected loss as defined in \eqref{eqn:org_objective} by applying the gradient descent method on compressor and critic iteratively. However, since there are discrete sampling operations, \ie dropout operations, within the computational graph, backpropagation is not directly applicable. Therefore we apply an unbiased likelihood-ratio estimator to calculate the gradient over $\phi$~\cite{glynn1990likelihood,peters2006policy}:
\begin{equation}
\small
\begin{split}
\nabla_{\phi} \mathcal{L} &= \sum_{\mathbf{z}} \nabla_{\phi} \mu_{\phi}(\mathcal{W}) \cdot L\big(\mathbf{y} , F_{\mathcal{W}}(\mathbf{x}|\mathbf{z})\big)\\
&= \sum_{\mathbf{z}} \mu_{\phi}(\mathcal{W}) \nabla_{\phi} \log \mu_{\phi}(\mathcal{W}) \cdot L\big(\mathbf{y} , F_{\mathcal{W}}(\mathbf{x}|\mathbf{z})\big)\\
&= \mathbb{E}_{\mathbf{z} \sim \mu_{\phi}} \big[\nabla_{\phi} \log \mu_{\phi}(\mathcal{W}) \cdot   L\big( \mathbf{y}, F_{\mathcal{W}}(\mathbf{x}|\mathbf{z})\big)  \big].
\end{split}
\label{eqn:compressor_grad}
\end{equation}

Therefore an unbiased estimator for \eqref{eqn:compressor_grad} can be
\begin{equation}
\small
\widehat{\nabla_{\phi} \mathcal{L}} = \nabla_{\phi} \log \mu_{\phi}(\mathcal{W}) \cdot   L\big( \mathbf{y}, F_{\mathcal{W}}(\mathbf{x}|\mathbf{z})\big)
\quad
\mathbf{z} \sim \mu_{\phi}.
\label{eqn:compressor_grad_mc}
\end{equation}

The gradient over $\mathbf{W}^{(l)} \in \mathcal{W}$ is
\begin{equation}
\small
\begin{split}
\nabla_{\mathbf{W}^{(l)}} \mathcal{L} &= \sum_{\mathbf{z}}  \mu_{\phi}(\mathcal{W}) \cdot \nabla_{\mathbf{W}^{(l)}} L\big(\mathbf{y} , F_{\mathcal{W}}(\mathbf{x}|\mathbf{z})\big)\\
&= \mathbb{E}_{\mathbf{z} \sim \mu_{\phi}} \big[  \nabla_{\mathbf{W}^{(l)}} L\big(\mathbf{y} , F_{\mathcal{W}}(\mathbf{x}|\mathbf{z})\big)  \big].
\end{split}
\label{eqn:predictor_grad}
\end{equation}

Similarly, an unbiased estimator for \eqref{eqn:predictor_grad} can be
\begin{equation}
\small
\widehat{\nabla_{\mathbf{W}^{(l)}} \mathcal{L}} =    \nabla_{\mathbf{W}^{(l)}} L\big( \mathbf{y}, F_{\mathcal{W}}(\mathbf{x}|\mathbf{z})\big)
\quad
\mathbf{z} \sim \mu_{\phi}.
\label{eqn:predictor_grad_mc}
\end{equation}

Now we provide more details of $\widehat{\nabla_{\phi} \mathcal{L}}$ in \eqref{eqn:compressor_grad_mc}. Although the estimator \eqref{eqn:compressor_grad_mc} is an unbiased estimator, it tends to have a higher variance. A higher variance of estimator can make the convergence slower. Therefore, variance reduction techniques are typically required to make the optimization feasible in practice~\cite{mnih2014neural, gu2015muprop}.

One variance reduction technique is to subtract a constant $c$ from learning signal $L\big(\mathbf{y} , F_{\mathcal{W}}(\mathbf{x}|\mathbf{z})\big)$ in \eqref{eqn:compressor_grad}, which still keeps the expectation of the gradient unchanged~\cite{mnih2014neural}.
Therefore, we keep track of the moving average of the learning signal $L\big(\mathbf{y} , F_{\mathcal{W}}(\mathbf{x}|\mathbf{z})\big)$ denoted by $c$, and subtract $c$ from the gradient estimator \eqref{eqn:compressor_grad_mc}.

The other variance reduction technique is keeping track of the moving average of the signal variance $v$, and divides the learning signal by $\max(1, \sqrt{v})$~\cite{gu2015muprop}.

Combing the aforementioned two variance reduction techniques, the final estimator \eqref{eqn:compressor_grad_mc} for gradient over $\phi$ becomes
\begin{equation}
\small
\widehat{\nabla_{\phi} \mathcal{L}} = \nabla_{\phi} \log \mu_{\phi}(\mathcal{W}) \cdot   \frac{L\big( \mathbf{y}, F_{\mathcal{W}}(\mathbf{x}|\mathbf{z})\big) - c}{\max(1, \sqrt{v})}
\quad
\mathbf{z} \sim \mu_{\phi},
\label{eqn:compressor_grad_mc_var_reduce}
\end{equation}
where $c$ and $v$ are the moving average of mean and the moving average of variance of learning signal $L\big(\mathbf{y} , F_{\mathcal{W}}(\mathbf{x}|\mathbf{z})\big)$ respectively.

After introducing the basic optimization process in DeepIoT, now we are ready to deliver the details of the compressing process.
Compared with previous compressing algorithms that gradually delete weights without rehabilitation~\cite{han2015deep}, DeepIoT applies ``soft" deletion by gradually suppressing the dropout probabilities of hidden elements with a decay factor $\gamma\in(0,1)$. During the experiments in Section~\ref{sec:evaluation}, we set $\gamma$ as the default value $0.5$. Since it is impossible to make the optimal compression decisions from the beginning, suppressing the dropout probabilities instead of deleting the hidden elements directly can provide the ``deleted" hidden elements changes to recover. This less aggressive compression process reduces the potential risk of irretrievable network damage and learning inefficiency.

During the compressing process, DeepIoT gradually increases the threshold of dropout probability $\tau$ from $0$ with step $\Delta$. The hidden elements with dropout probability, $\mathbf{p}_{[j]}^{(l)}$ that is less than the threshold $\tau$ will be given decay on dropout probability, \ie $\hat{\mathbf{p}}_{[j]}^{(l)} \leftarrow \gamma \cdot \mathbf{p}_{[j]}^{(l)}$. Therefore, the operation in compressor (\ref{eqn:compressor}) can be updated as
\begin{equation}
\small
\mathbf{z}_{[j]}^{(l)} \sim \text{Bernoulli}\Big(\mathbf{p}_{[j]}^{(l)}\cdot \gamma^{\mathbbm{1}{\mathbf{p}_{[j]}^{(l)} \le \tau}}\Big),
\label{eqn:compressor_update}
\end{equation}

\begin{algorithm}        
\footnotesize
\caption{Compressor-predictor compressing process}   
\label{alg:comp_pred}                     
\begin{algorithmic}[1]                
\STATE \textbf{Input:} pre-trained predictor $F_{\mathcal{W}}(\mathbf{x}|\mathbf{z})$ 
\STATE \textbf{Initialize:} compressor $\mu_{\phi}(\mathcal{W})$ with parameter $\phi$, moving average $c$, moving average of variance $v$
\WHILE{$\mu_{\phi}(\mathcal{W})$ is not convergent}
\STATE $\mathbf{z}\sim \mu_\phi(\mathcal{W})$
\STATE $c \leftarrow \text{movingAvg}\big(L\big( \mathbf{y}, F_{\mathcal{W}}(\mathbf{x}|\mathbf{z})\big)\big)$
\STATE $v \leftarrow \text{movingVar}\big(L\big( \mathbf{y}, F_{\mathcal{W}}(\mathbf{x}|\mathbf{z})\big)\big)$
\STATE $ \phi \leftarrow \phi - \beta \cdot\nabla_{\phi} \log \mu_{\phi}(\mathcal{W}) \cdot   \big(L\big( \mathbf{y}, F_{\mathcal{W}}(\mathbf{x}|\mathbf{z})\big) - c\big)/{\max(1, \sqrt{v})}$
\ENDWHILE
\STATE $\tau = 0$
\WHILE{the percentage of left number of parameters in $F_{\mathcal{W}}(\mathbf{x}|\mathbf{z})$ is larger than $\alpha$}
\STATE $\mathbf{z}\sim \mu_\phi(\mathcal{W})$
\STATE $c \leftarrow \text{movingAvg}\big(L\big( \mathbf{y}, F_{\mathcal{W}}(\mathbf{x}|\mathbf{z})\big)\big)$
\STATE $v \leftarrow \text{movingVar}\big(L\big( \mathbf{y}, F_{\mathcal{W}}(\mathbf{x}|\mathbf{z})\big)\big)$
\STATE $ \phi \leftarrow \phi - \beta \cdot\nabla_{\phi} \log \mu_{\phi}(\mathcal{W}) \cdot   \big(L\big( \mathbf{y}, F_{\mathcal{W}}(\mathbf{x}|\mathbf{z})\big) - c\big)/{\max(1, \sqrt{v})}$
\STATE $\mathcal{W} \leftarrow \mathcal{W} - \beta  \cdot  \nabla_{\mathcal{W}} L\big( \mathbf{y}, F_{\mathcal{W}}(\mathbf{x}|\mathbf{z})\big)$
\STATE   update threshold $\tau$: $\tau \leftarrow \tau +\Delta$ for every $T$ rounds
\ENDWHILE
\STATE $\mathbf{\hat{z}}_{[j]}^{(l)} = {\mathbbm{1}{\mathbf{p}_{[j]}^{(l)} > \tau}}$
\WHILE {$F_{\mathcal{W}}(\mathbf{x}|\mathbf{\hat{z}})$ is not convergent}
\STATE $\mathcal{W} \leftarrow \mathcal{W} - \beta \cdot  \nabla_{\mathcal{W}} L\big( \mathbf{y}, F_{\mathcal{W}}(\mathbf{x}|\mathbf{\hat{z}})\big)$
\ENDWHILE
 \end{algorithmic}
\end{algorithm}

\noindent where $\mathbbm{1}$ is the indicator function; $\gamma \in (0,1)$ is the decay factor; and $\tau \in [0,1)$ is the threshold. Since the operation of suppressing dropout probability with the pre-defined decay factor $\gamma$ is differentiable, we can still optimize the original and the compressor neural network through \eqref{eqn:predictor_grad_mc} and \eqref{eqn:compressor_grad_mc_var_reduce}. The compression process will stop when the percentage of left number of parameters in $F_{\mathcal{W}}(\mathbf{x}|\mathbf{z})$ is smaller than a user-defined value $\alpha \in (0,1)$.

After the compression, DeepIoT fine-tunes the compressed model $F_{\mathcal{W}}(\mathbf{x}|\mathbf{\hat{z}})$, with a fixed mask $\mathbf{\hat{z}}$, which is decided by the previous threshold $\tau$. Therefore the mask generation step in \eqref{eqn:compressor_update} will be updated as
\begin{equation}
\small
\mathbf{\hat{z}}_{[j]}^{(l)} = {\mathbbm{1}{\mathbf{p}_{[j]}^{(l)} > \tau}}.
\label{eqn:final_mask}
\end{equation}

We summarize the compressor-critic compressing process of DeepIoT in Algorithm~\ref{alg:comp_pred}.

The algorithm consists of three parts. In the first part (Line 3 to Line 8), DeepIoT freezes the critic $F_{\mathcal{W}}(\mathbf{x}|\mathbf{z})$ and initializes the compressor $\mu_{\phi}(\mathcal{W})$ according to \eqref{eqn:compressor_grad_mc_var_reduce}. In the second part (Line 9 to Line 17), DeepIoT optimizes the critic and compressor jointly with the gradients calculated by \eqref{eqn:predictor_grad_mc} and \eqref{eqn:compressor_grad_mc_var_reduce}. At the same time, DeepIoT gradually compresses the predictor by suppressing dropout probabilities according to \eqref{eqn:compressor_update}. In the final part (Line 18 to Line 21), DeepIoT fine-tunes the critic with the gradient calculated by \eqref{eqn:predictor_grad_mc} and a deterministic dropout mask is generated according to  \eqref{eqn:final_mask}. After these three phases, DeepIoT generates a binary dropout mask $\mathbf{\hat{z}}$ and the fine-tuning parameters of the critic $\mathcal{W}$. With these two results, we can easily obtain the compressed model of the original neural network.

\section{Implementation}~\label{sec:implementation}
In this section, we briefly describe the hardware, software, architecture, and performance summary of DeepIoT.

\vspace{-0.1cm}
\subsection{Hardware}
Our hardware is based on Intel Edison computing platform~\cite{Edison}.
The Intel Edison computing platform is powered by the Intel Atom SoC dual-core CPU at 500 MHz and is equipped with 1GB memory and 4GB flash storage. For fairness, all neural network models are run solely on CPU during experiments.

\vspace{-0.1cm}
\subsection{Software}
All the original neural networks for all sensing applications mentioned in Section~\ref{sec:evaluation} are trained on the workstation with NVIDIA GeForce GTX Titan X. For all baseline algorithms mentioned in Section~\ref{sec:evaluation}, the compressing processes are also conducted on the workstation. The compressed models are exported and loaded into the flash storage on Intel Edison for experiments.

We installed the Ubilinux operation system on Intel Edison computing platform~\cite{ubilinux}. Far fairness, all compressed deep learning models are run through Theano~\cite{2016arXiv160502688short} with only CPU device on Intel Edison. The matrix multiplication operations and sparse matrix multiplication operations are optimized by BLAS and Sparse BLAS respectively during the implementation. No additional run-time optimization is applied for any compressed model and in all experiments. 

\vspace{-0.1cm}
\subsection{Architecture}
Given the original neural network structure and parameters as well as the device resource information, DeepIoT can automatically generate a compressed neural network that is ready to be run on embedded devices with sensor inputs. The system first obtains the memory information from the embedded device and sets the final compressed size of the neural network to fit in a pre-configured fraction of available memory, from which the needed compression ratio is computed. In the experiments, we manually set the ratio to exploit the capability of DeepIoT. This ratio, together with the parameters of the original model are then used to automatically generate the corresponding compressor neural network to compress the original neural network. The resulting compressed neural network is transferred to the embedded device. This model can be then called locally with a data input to decide on the output. The semantics of input and output are not known to the model.

\subsection{Performance Summary}
We list the resource consumption numbers of all compressed models without loss of accuracy generated by DeepIoT and their corresponding original model in Table~\ref{tab:overallConsum} with the form of (original/compressed/reduction percentage). These models are explained in more detail in the evaluation, Section~\ref{sec:evaluation}.

\begin{table}[!htb]
\footnotesize
\begin{center}
\caption {Resource consumptions of model implementations on Intel Edison}
\vspace{-0.1cm}
\label{tab:overallConsum}
\begin{tabular}{ |c | c | c | c | } 
 \hline
 Model & Size (MB) & Time (ms) & Energy (mJ) \\ 
  \hline
   \hline
  LeNet5 & $1.72/0.04/97.6\%$  & $50.2/14.2/71.4\%$ & $47.1/12.5/73.5\%$ \\ 
  \hline
  VGGNet & $118.8/2.9/97.6\%$ & $1.5K/82.2/94.5\%$ & $1.7K/74/95.6\%$ \\ 
 \hline
 Bi-LSTM  & $76.0/7.59/90.0\%$ & $71K/9.6K/86.5\%$ & $62.9K/8.1K/87.1\%$ \\ 
 \hline
 DeepSense1  & $1.89/0.12/93.7\%$ & $130/36.7/71.8\%$ & $99.6/27.7/72.2\%$ \\ 
 \hline
 DeepSense2  & $1.89/0.02/98.9\%$ & $130/25.1/80.7\%$ & $105.1/18.1/82.8\%$ \\ 
 \hline
\end{tabular}
\end{center}
\end{table}

Although the models generated by DeepIoT do not use sparse matrix representations, other baseline algorithms, as will be introduced in Section~\ref{sec:evaluation}, may use sparse matrices to represent models. When the proportion of non-zero elements in the sparse matrix is larger than $20\%$, sparse matrix multiplications can even run slower than their non-sparse counterpart. Therefore, there is a tradeoff between memory consumption and execution time for sparse matrices with a large proportion of non-zero elements. In addition, convolution operations conducted on CPU are also formulated and optimized as matrix multiplications, as mentioned in Section~\ref{sec:dropout}. Therefore, the tradeoff still exists. For all baseline algorithms in Section~\ref{sec:evaluation}, we implement both the sparse matrix version and the non-sparse matrix version. During all the experiments with baseline algorithms, we ``cheat", in their favor, by choosing the version that performs better according to the current evaluation metrics.

\section{Evaluation}~\label{sec:evaluation}
In this section, we evaluate DeepIoT through two sets of experiments. 
The first set is motivated by the prospect of enabling future smarter embedded ``things" (physical objects) to interact with humans using user-friendly modalities such as visual cues, handwritten text, and speech commands, while the second evaluates human-centric context sensing, such as human activity recognition and user identification. In the following subsections, we first describe the comparison baselines that are current state of the art deep neural network compression techniques. We then present the first set of experiments that demonstrate accuracy and resource demands observed if IoT-style smart objects  
interacted with users via natural human-centric modalities thanks to deep neural networks compressed, for the resource-constrained hardware, with the help of our DeepIoT framework.
Finally, we present the second set of experiments that demonstrate accurancy and resource demands when applying DeepIoT to compress deep neural networks trained for human-centric context sensing applications. In both cases, we show significant advantages in the accuracy/resource trade-off over the compared state-of-the-art compression baselines.

\subsection{Baseline Algorithms}
We compare DeepIoT with other three baseline algorithms:
\begin{enumerate}[leftmargin=*]
\setlength\itemsep{0pt}
\item \textbf{DyNS:} This is a magnitude-based network pruning algorithm~\cite{guo2016dynamic}. The algorithm prunes weights in convolutional kernels and fully-connected layer based on the magnitude. It retrains the network connections after each pruning step and has the ability to recover the pruned weights. For convolutional and fully-connected layers, DyNS searches the optimal thresholds separately.
\item \textbf{SparseSep:} This is a sparse-coding and factorization based algorithm~\cite{lane2016sparsifying}. The algorithm simplifies the fully-connected layer by finding the optimal code-book and code based on a sparse coding technique. For the convolutional layer, the algorithm compresses the model with matrix factorization methods. We greedily search for the optimal code-book and factorizaiton number from the bottom to the top layer.
\item \textbf{DyNS-Ext:} The previous two algorithms mainly focus on compressing convolutional and fully-connected layers. Therefore we further enhance and extend the magnitude-based method used in DyNS to recurrent layers and call this algorithm DyNS-Ext. Just like DeepIoT, DyNS-Ext can be applied to all commonly used deep network modules, including fully-connected layers, convolutional layers, and recurrent layers. If the network structure does not contain recurrent layers, we apply DyNS instead of DyNS-Ext.
\end{enumerate}
For magnitude-based pruning algorithms, DyNS and DyNS-Ext, hidden elements with zero input connections or zero output connections will be pruned to further compress the network structure.
In addition, all models use 32-bit floats without any quantization.

\subsection{Supporting Human-Centric Interaction Modalities}
Three basic interaction modalities among humans are text, vision, and speech. In this section, we describe three different experiments that test implementations of these basic interaction modalities on low-end devices using trained and compressed neural networks. 
We train state-of-art neural networks on traditional benchmark datasets as original models. Then, we compress the original models using DeepIoT and the three baseline algorithms described above. Finally, we test the accuracy and resource consumption that result from using these compressed models on the embedded device.

\subsubsection{Handwritten digits recognition with LeNet5}
The first human interaction modality is recognizing handwritten text. In this experiment, we consider a meaningful subset of that; namely recognizing handwritten digits from visual inputs. An example application that uses this capability might be a smart wallet equipped with a camera and a tip calculator. 
We use MNIST\footnote{http://yann.lecun.com/exdb/mnist/} as our training and testing dataset.
The MNIST is a dataset of handwritten digits that is commonly used for training various image processing systems. It has a training set of $60000$ examples, and a test set of $10000$ examples.

We test our algorithms and baselines on the LeNet-5 neural network model. The corresponding network structure is shown in Table~\ref{tab:MNIST}. Notice that we omit all the polling layers in Table~\ref{tab:MNIST} for simplicity, because they do not contain training parameters.

\begin{table*}[!htb]
\footnotesize
\begin{center}
\caption {LeNet5 on MNIST dataset}
\label{tab:MNIST}
\begin{tabular}{ |c| c | c | c | c |c | c| } 
 \hline
 Layer & Hidden Units & Params & \multicolumn{2}{ c| }{DeepIoT (Hidden Units/ Params)} & DyNS & SparseSep\\ 
  \hline
   \hline
  conv1 $(5\times 5)$ & $20$  & $0.5$K & $10$ & $50.0\%$ & $24.2\%$ & $84\%$ \\ 
  \hline
  conv2 $(5\times 5)$  & $50$ & $25$K & $20$ & $20.0\%$ & $20.7\%$ & $91\%$ \\ 
 \hline
 fc1  & $500$ & $400$K & $10$  & $0.8\%$ & $1.0\%$ & $78.75\%$ \\ 
 \hline
 fc2  & $10$ & $5$K & $10$ & $2.0$\%  & $16.34\%$ & $70.28\%$ \\ 
 \hline
 total  & \diagbox[dir=SW,width=2.2cm, height=0.31cm]{}{} & $431$K & \diagbox[dir=SW,width=2.2cm, height=0.31cm]{}{}  & $1.98\%$ & $2.35\%$ & $72.39\%$ \\ 
 \hline
 \hline
  Test Error & \multicolumn{2}{ c| }{$0.85\%$}   &  \multicolumn{2}{ c| }{$0.85\%$} & $0.85\%$ & $1.05\%$\\ 
 \hline
\end{tabular}
\end{center}
\end{table*}

\begin{figure*}[!htb]
\begin{subfigure}{.32\linewidth}
  \centering
  \includegraphics[width=1.\linewidth]{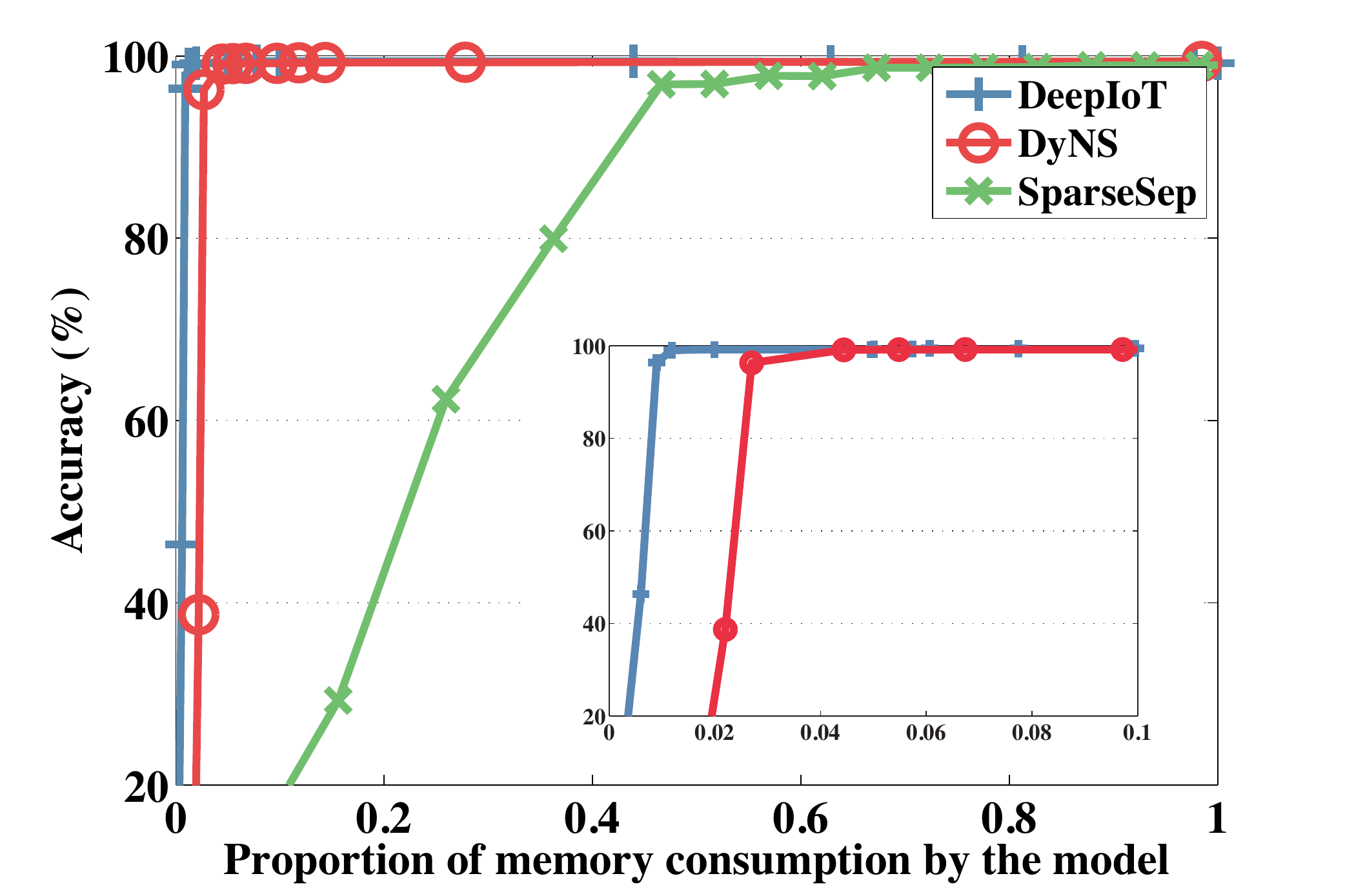}
  \caption{The tradeoff between testing accuracy and memory consumption by models. }
  \label{fig:lenet5_acc_mem}
\end{subfigure}%
\begin{subfigure}{.32\linewidth}
  \centering
  \includegraphics[width=1.\linewidth]{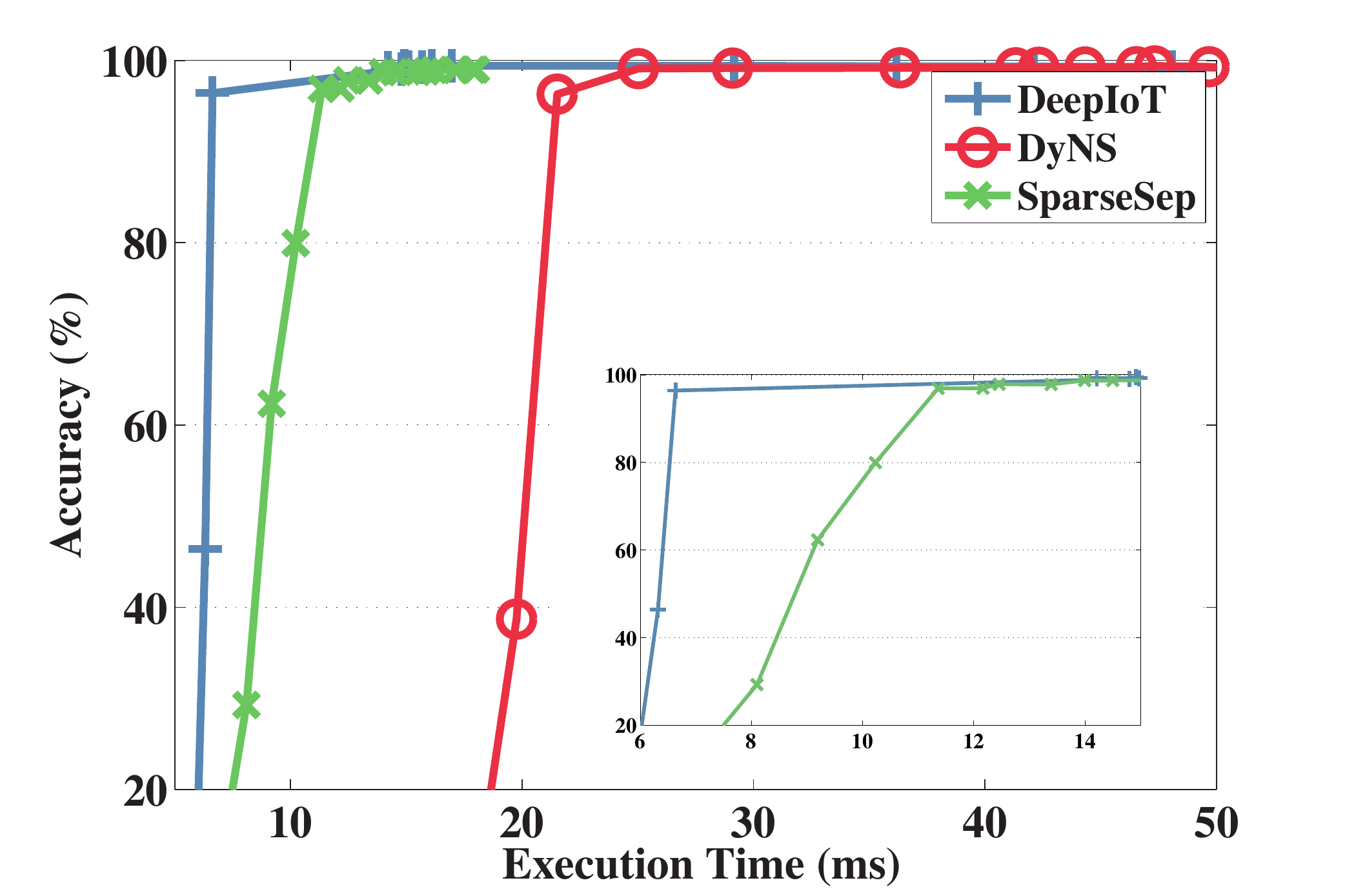}
  \caption{The tradeoff between testing accuracy and execution time.}
  \label{fig:lenet5_acc_time}
\end{subfigure}
\begin{subfigure}{.32\linewidth}
  \centering
  \includegraphics[width=1.\linewidth]{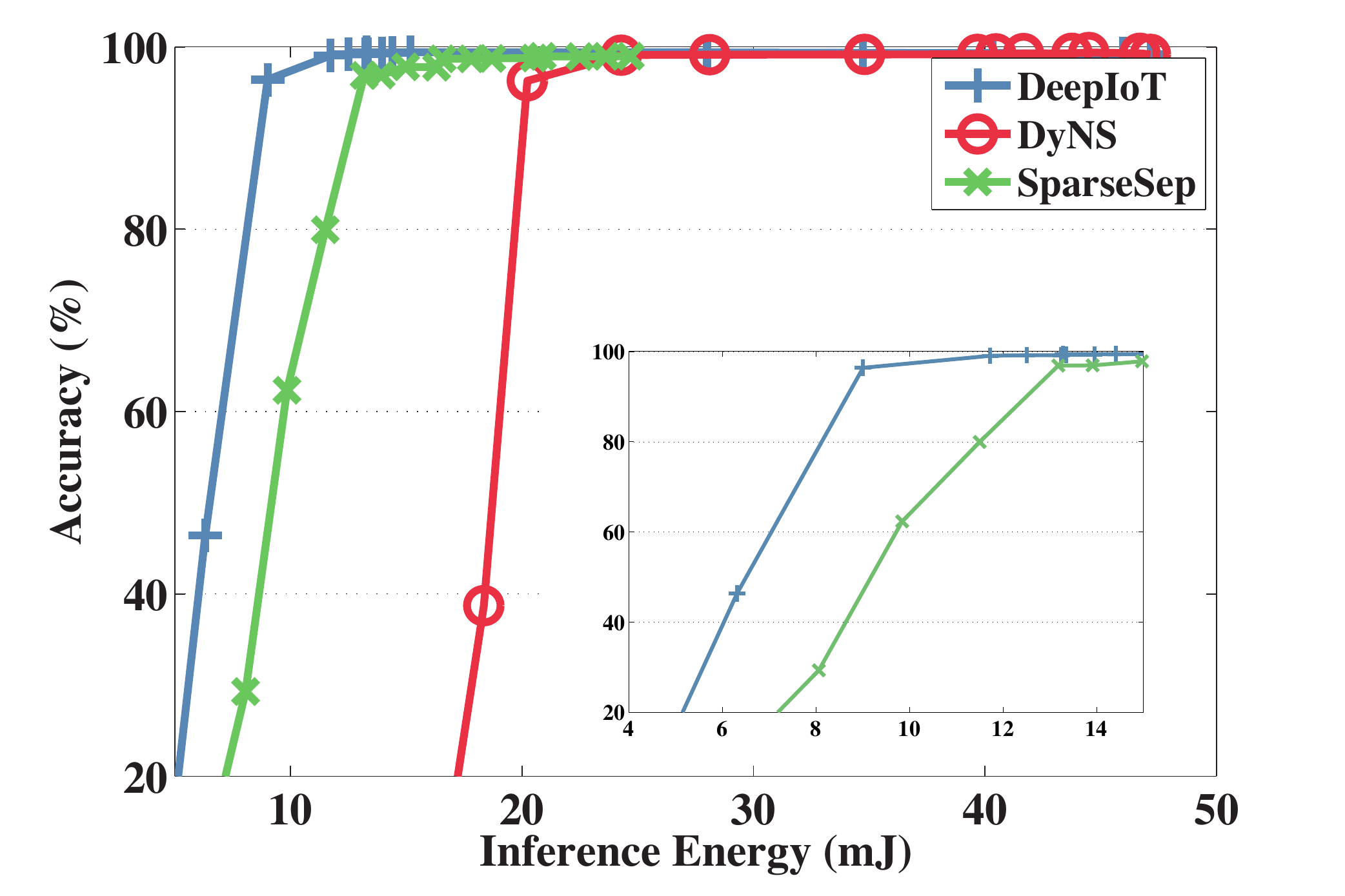}
  \caption{The tradeoff between testing accuracy and energy consumption.}
  \label{fig:lenet5_acc_engy}
\end{subfigure}
\caption{System performance tradeoff for LeNet5 on MNIST dataset}
\label{fig:letnet5}
\end{figure*}

The first column of Table~\ref{tab:MNIST} represents the network structure of LeNet-5, where ``convX" represents the convolutional layer and ``fcY" represents the fully-connected layer. The second column represents the number of hidden units or convolutional kernels we used in each layer. The third column represents the number of parameters used in each layer and in total. The original LeNet-5 is trained and achieves an error rate of $0.85\%$ in the test dataset. 

We then apply DeepIoT and two other baseline algorithms, DyNS and SparseSep, to compress LeNet-5. Note that, we do not use DyNS-Ext because the network does not contain a recurrent layer. The network statistics of the compressed model are shown in Table~\ref{tab:MNIST}. DeepIoT is designed to prune the number of hidden units for a more efficient network structure. Therefore, we illustrate both the remaining number of hidden units and the proportion of the remaining number of parameters in Table~\ref{tab:MNIST}. Both DeepIoT and DyNS can significantly compress the network without hurting the final performance. SparseSep shows an acceptable drop of performance. This is because SparseSep is designed without fine-tuning. It has the benefit of not fine-tuning the model, but it suffers the loss in the final performance at the same time.

The detailed tradeoff between testing accuracy and memory consumption by the model is illustrated in Fig~\ref{fig:lenet5_acc_mem}. We compress the original neural network with different compression ratios and recode the final testing accuracy. In the zoom-in illustration, DeepIoT achieves at least $\times 2$ better tradeoff compared with the two baseline methods.  This is mainly due to two reasons. One is that the compressor neural network in DeepIoT 
obtains a global view of parameter redundancies and is therefore better capable of eliminating them.
The other is that DeepIoT prunes the hidden units directly, which enables us to represent the compressed model parameters with a small dense matrix instead of a large sparse matrix. The sparse matrix consumes more memory for the indices of matrix elements. Algorithms such as DyNS generate models represented by sparse matrices that cause larger memory consumption.

The evaluation results on execution time of compressed models on Intel Edison, are illustrated in Fig.~\ref{fig:lenet5_acc_time}. We run each compressed model on Intel Edison for 5000 times and use the mean value for generating the tradeoff curves.

DeepIoT still achieves the best tradeoff compared with other two baselines by a significant margin. DeepIoT takes $14.2$ms to make a single inference, which reduces execution time by $71.4\%$ compared with the original network without loss of accuracy. However SparseSep takes less execution time compared with DyNS at the cost of acceptable performance degradation (around 0.2\% degradation on test error). The main reason for this observation is that, even though fully-connected layers occupy the most model parameters, most execution time is used by the convolution operations. SparseSep uses a matrix factorization method to covert the 2d convolutional kernel into two 1d convolutional kernels on two different dimensions~\cite{tai2015convolutional}. Although this method makes low-rank assumption on convolutional kernel, it can speed up convolution operations if the size of convolutional kernel is large ($5\times 5$ in this experiment). It can sometimes speed up the operation even when two 1d kernels have more parameters in total compared with the original 2d kernel. However DyNS applies a magnitude-based method that prunes most of the parameters in fully-connected layers. For convolutional layers, DyNS does not reduce the number of convolutional operations effectively, and sparse matrix multiplication is less efficient compared with regular matrix with the same number of elements. DeepIoT directly reduces the number of convolutional kernels in each layer, which reduces the number of operations in convolutional layers without making the low-rank assumption that can hurt the network performance.

The evaluation of energy consumption on Intel Edison is illustrated in Fig.~\ref{fig:lenet5_acc_engy}. For each compressed model, we run it for 5000 times and measure the total energy consumption by a power meter. Then, we calculate the expected energy consumption for one-time execution and use the one-time energy consumption to generate the tradeoff curves in Fig.~\ref{fig:lenet5_acc_engy}.

Not surprisingly, DeepIoT still achieves the best tradeoff in the evaluation on energy consumption by a significant margin. It reduces energy consumption by $73.7\%$  compared with the original network without loss of accuracy. Being similar as the evaluation on execution time, energy consumption focuses more on the number of operations than the model size. Therefore, SparseSep can take less energy consumption compared with DyNS at the cost of acceptable loss on performance.

\begin{table*}[!htb]
\footnotesize
\begin{center}
\caption {VGGNet on CIFAR-10 dataset}
\label{tab:VGGNet}
\vspace{-0.3cm}
\begin{tabular}{ |c| c | c | c | c |c | c| } 
 \hline
 Layer & Hidden Units & Params & \multicolumn{2}{ c| }{DeepIoT (Hidden Units/ Params)} & DyNS & SparseSep\\ 
  \hline
   \hline
  conv1 $(3\times 3)$ & $64$  & $1.7$K & $27$ & $42.2\%$ & $53.9\%$ & $93.1\%$ \\ 
  \hline
  conv2 $(3\times 3)$  & $64$ & $36.9$K & $47$ & $31.0\%$ & $40.1\%$ & $57.3\%$ \\ 
 \hline
  conv3 $(3\times 3)$  & $128$ & $73.7$K & $53$ & $30.4\%$ & $52.3\%$ & $85.1\%$ \\ 
 \hline
  conv4 $(3\times 3)$  & $128$ & $147.5$K & $68$ & $22.0\%$ & $67.0\%$ & $56.8\%$ \\ 
 \hline
  conv5 $(3\times 3)$  & $256$ & $294.9$K & $104$ & $21.6\%$ & $71.2\%$ & $85.1\%$ \\ 
 \hline
  conv6 $(3\times 3)$  & $256$ & $589.8$K & $97$ & $15.4\%$ & $65.0\%$ & $56.8\%$ \\ 
 \hline
 conv7 $(3\times 3)$  & $256$ & $589.8$K & $89$ & $13.2\%$ & $61.2\%$ & $56.8\%$ \\ 
 \hline
  conv8 $(3\times 3)$  & $512$ & $1.179$M & $122$ & $8.3\%$ & $36.5\%$ & $85.2\%$ \\ 
 \hline
 conv9 $(3\times 3)$  & $512$ & $2.359$M & $95$ & $4.4\%$ & $10.6\%$ & $56.8\%$ \\ 
 \hline
 conv10 $(3\times 3)$  & $512$ & $2.359$M & $64$ & $2.3\%$ & $3.9\%$ & $56.8\%$ \\ 
 \hline
 conv11 $(2\times 2)$  & $512$ & $1.049$M & $128$ & $3.1\%$ & $3.0\%$ & $85.2\%$ \\ 
 \hline
 conv12 $(2\times 2)$  & $512$ & $1.049$M & $112$ & $5.5\%$ & $1.7\%$ & $85.2\%$ \\ 
 \hline
 conv13 $(2\times 2)$  & $512$ & $1.049$M & $149$ & $6.4\%$ & $2.4\%$ & $85.2\%$ \\ 
 \hline
 fc1  & $4096$ & $2.097$M & $27$  & $0.19\%$ & $2.2\%$ & $95.8\%$ \\ 
 \hline
 fc2  & $4096$ & $16.777$M & $371$  & $0.06\%$ & $0.39\%$ & $135\%$ \\ 
 \hline
 fc3  & $10$ & $41$K & $10$ & $9.1$\%  & $18.5\%$ & $90.2\%$ \\ 
 \hline
 total  & \diagbox[dir=SW,width=2.2cm, height=0.31cm]{}{} & $29.7$M & \diagbox[dir=SW,width=2.2cm, height=0.31cm]{}{}  & $2.44\%$ & $7.05\%$ & $112\%$ \\ 
 \hline
 \hline
  Test Accuracy & \multicolumn{2}{ c| }{$90.6\%$}   &  \multicolumn{2}{ c| }{$90.6\%$} & $90.6\%$ & $87.1\%$\\ 
 \hline
\end{tabular}
\end{center}
\end{table*}

\begin{figure*}[!htb]
\vspace{-0.3cm}
\begin{subfigure}{.32\linewidth}
  \centering
  \includegraphics[width=1.\linewidth]{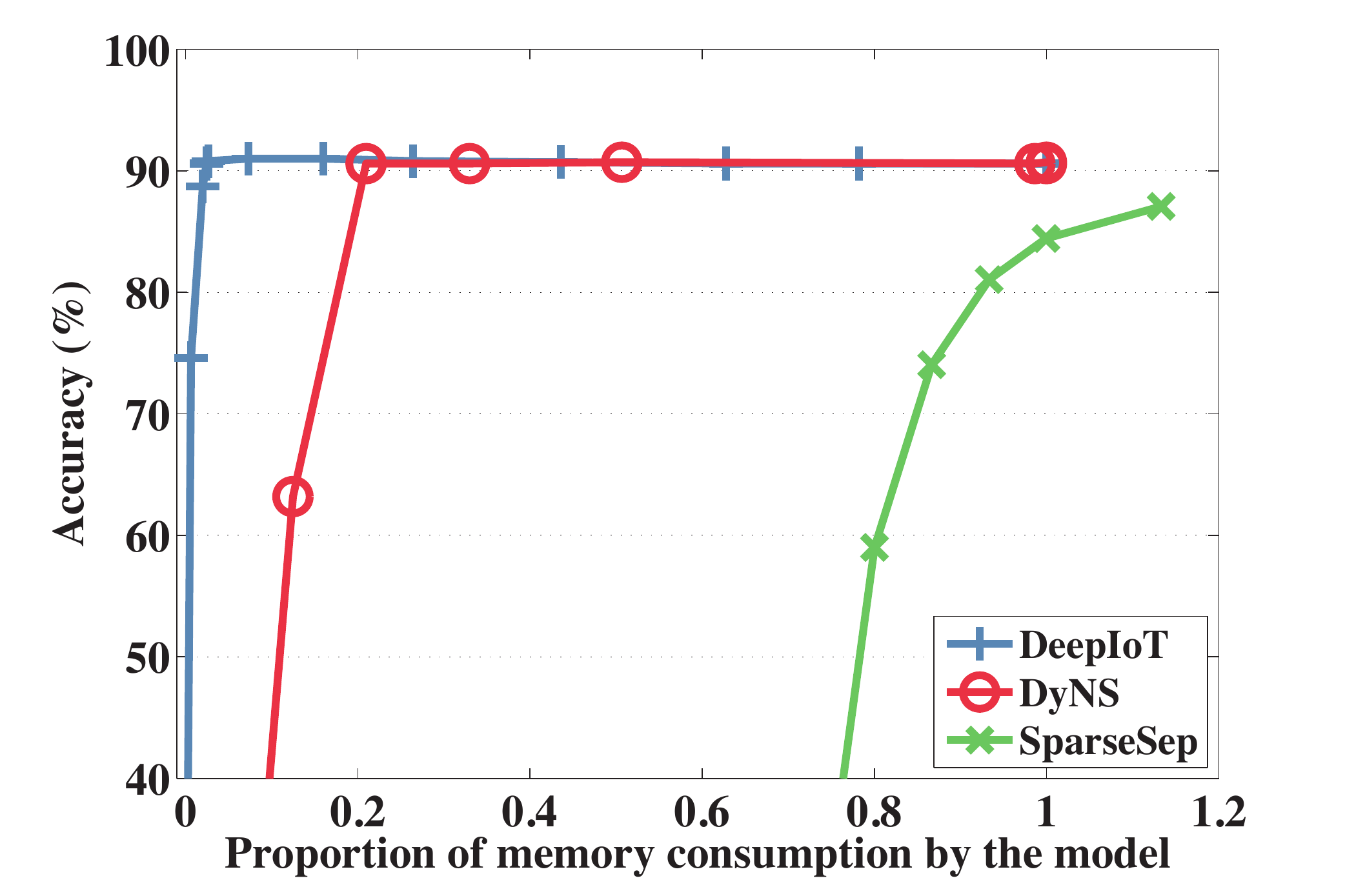}
  \caption{The tradeoff between testing accuracy and memory consumption by models. }
  \label{fig:vggnet_acc_mem}
\end{subfigure}%
\begin{subfigure}{.32\linewidth}
  \centering
  \includegraphics[width=1.\linewidth]{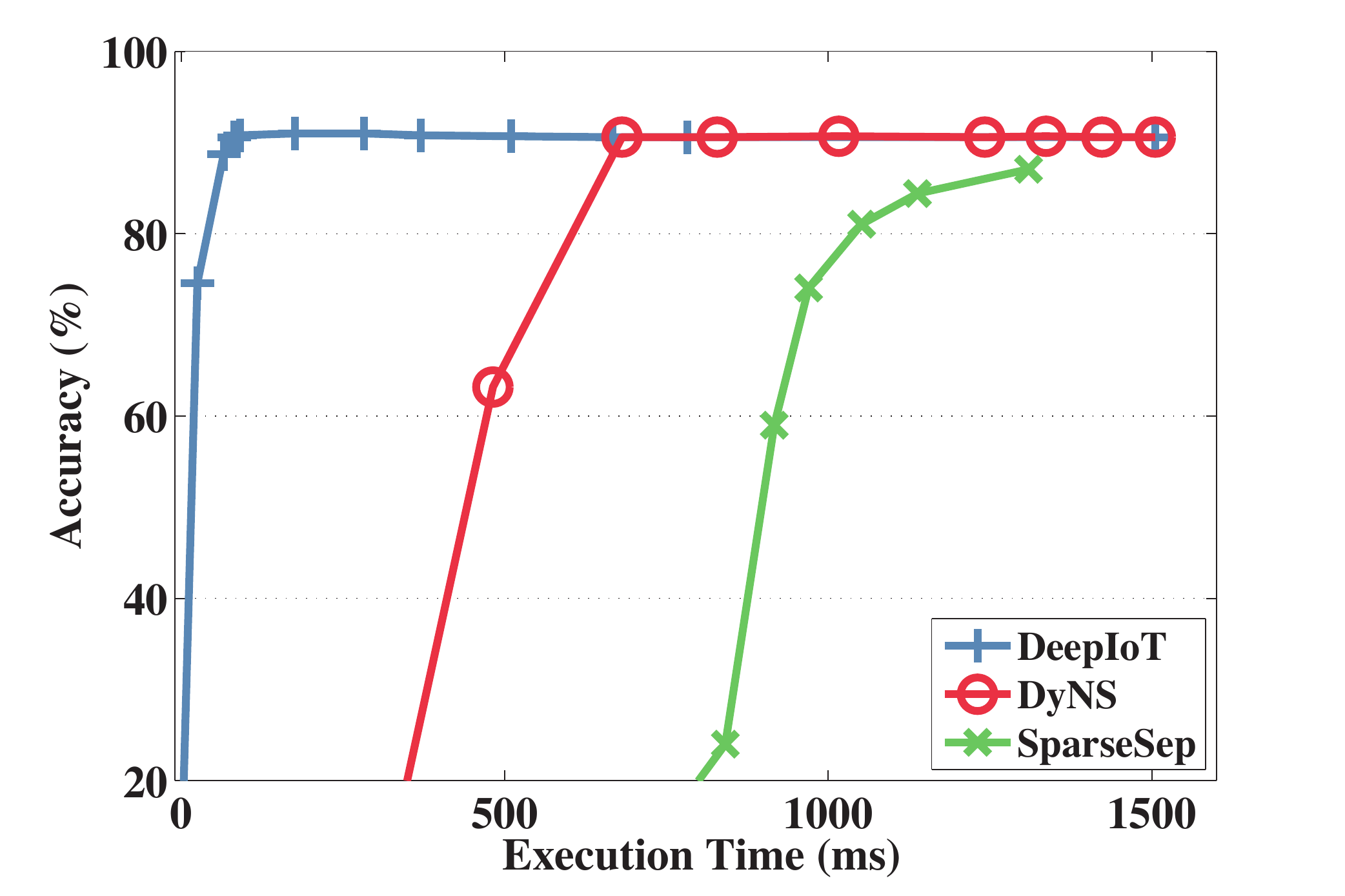}
  \caption{The tradeoff between testing accuracy and execution time.}
  \label{fig:vggnet_acc_time}
\end{subfigure}
\begin{subfigure}{.32\linewidth}
  \centering
  \includegraphics[width=1.\linewidth]{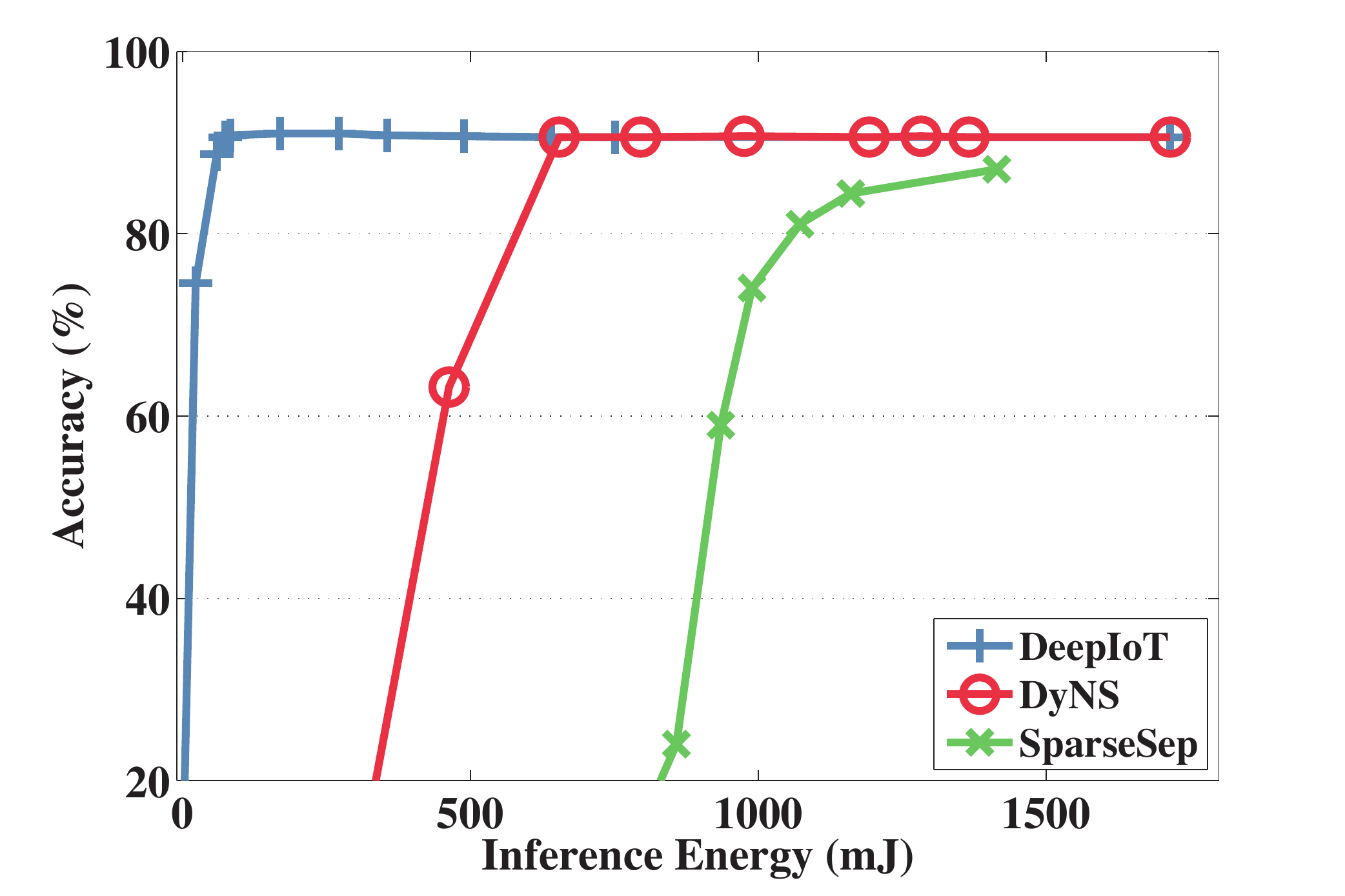}
  \caption{The tradeoff between testing accuracy and energy consumption.}
  \label{fig:vggnet_acc_engy}
\end{subfigure}
\caption{System performance tradeoff for VGGNet on CIFAR-10 dataset}
\label{fig:vggnet}
\vspace{-0.35cm}
\end{figure*}

\subsubsection{Image recognition with VGGNet}
The second human interaction modality is through vision.
During this experiment, we use CIFAR10\footnote{https://www.kaggle.com/c/cifar-10} as our training and testing dataset.
The CIFAR-10 dataset consists of 60000 32x32 colour images in 10 classes, with 6000 images per class. There are 50000 training images and 10000 test images. It is a standard testing benchmark dataset for the image recognition tasks. While not necessarily representative of seeing objects in the wild, it offers a more controlled environment for an apples-to-apples comparison.    

During this evaluation, we use the VGGNet structure as our original network structure. It is a huge network with millions of parameters. VGGNet is chosen to show that DeepIoT is able to compress relative deep and large network structure. The detailed structure is shown Table~\ref{tab:VGGNet}.

In Table~\ref{tab:VGGNet}, we illustrate the detailed statistics of best compressed model that keeps the original testing accuracy for three algorithms. We clearly see that DeepIoT beats the other two baseline algorithms by a significant margin. This shows that the compressor in DeepIoT can handle networks with relatively deep structure. The compressor uses a variant of the LSTM architecture to share the redundancy information among different layers. Compared with other baselines considering only local information within each layer, sharing the global information among layers helps us learn about the parameter redundancy and compress the network structure. In addition, we observe performance loss in the compressed network generated by SparseSep. It is mainly due to the fact that SparseSep avoids the fine-tuning step. This experiment shows that fine-tuning (Line 18 to Line 21 in Algorithm~\ref{alg:comp_pred}) is important for model compression.

Fig.~\ref{fig:vggnet_acc_mem} shows the tradeoff between testing accuracy and memory consumption for different models. DeepIoT 
achieves a better performance by even a larger margin,
because the 
model generated by DeepIoT can still be represented by a standard matrix, while other methods that use a sparse matrix representation require more memory consumption.

\begin{table*}[!htb]
\footnotesize
\begin{center}
\caption {Deep bidirectional LSTM on LibriSpeech ASR corpus}
\label{tab:LSTM}
\vspace{-0.3cm}
\begin{tabular}{ |c| c | c | c | c | c | c | c | c | c | c | c |} 
 \hline
 \multicolumn{2}{ |c| }{Layer} & \multicolumn{2}{ c| }{Hidden Unit} & \multicolumn{2}{ c| }{Params} & \multicolumn{4}{ c| }{DeepIoT (Hidden Units/ Params)} & \multicolumn{2}{ c| }{DyNS-Ext}  \\ 
  \hline
   \hline
  $\phantom{a}$LSTMf1$\phantom{a}$ & LSTMb1 & $\phantom{\tiny^1}512\phantom{\tiny^1}$ & $512$ & 1.090M & 1.090M & 55 & 20 & $\phantom{1}$10.74\%$\phantom{1}$ & 3.91\% & 34.9\% & 18.2\%\\ 
  \hline
  LSTMf2 & LSTMb2 & 512 & 512 & 2.097M & 2.097M & 192 & 71 & 4.03\% & 0.54\% & 37.2\% & 23.1\%  \\ 
  \hline
  LSTMf3 & LSTMb3 & 512 & 512 & 2.097M & 2.097M & 240 & 76 & 17.58\%& 2.06\% & 43.1\% & 27.9\%  \\ 
  \hline
   LSTMf4 & LSTMb4 & 512 & 512 & 2.097M & 2.097M & 258 & 81 & 23.62\%& 2.35\% & 52.3\% & 40.2\% \\ 
  \hline
  LSTMf5 & LSTMb5 & 512 & 512 & 2.097M & 2.097M & 294 & 90 & 28.93\% & 2.78\% & 72.6\% & 61.8\% \\ 
    \hline
  \multicolumn{2}{ |c| }{fc1 } & \multicolumn{2}{ c| }{29}& \multicolumn{2}{ c| }{59.3K} & \multicolumn{2}{ c| }{29} & \multicolumn{2}{ c| }{37.5\%}& \multicolumn{2}{ c| }{69.0\%}\\ 
  \hline
   \multicolumn{2}{ |c| }{total } & \multicolumn{2}{ c| }{\diagbox[dir=SW,width=2.cm, height=0.31cm]{}{}}& \multicolumn{2}{ c| }{19.016M} & \multicolumn{2}{ c| }{\diagbox[dir=SW,width=1.3cm, height=0.31cm]{}{}} & \multicolumn{2}{ c| }{9.98\%}& \multicolumn{2}{ c| }{37.1\%}\\ 
  \hline
  \hline
   \multicolumn{2}{ |c| }{Word error rate (WER)} & \multicolumn{4}{ c| }{9.31} & \multicolumn{4}{ c| }{9.20} & \multicolumn{2}{ c| }{9.62} \\
    \hline
\end{tabular}
\end{center}
\end{table*}

\begin{figure*}[!htb]
\vspace{-0.3cm}
\begin{subfigure}{.32\linewidth}
  \centering
  \includegraphics[width=1.\linewidth]{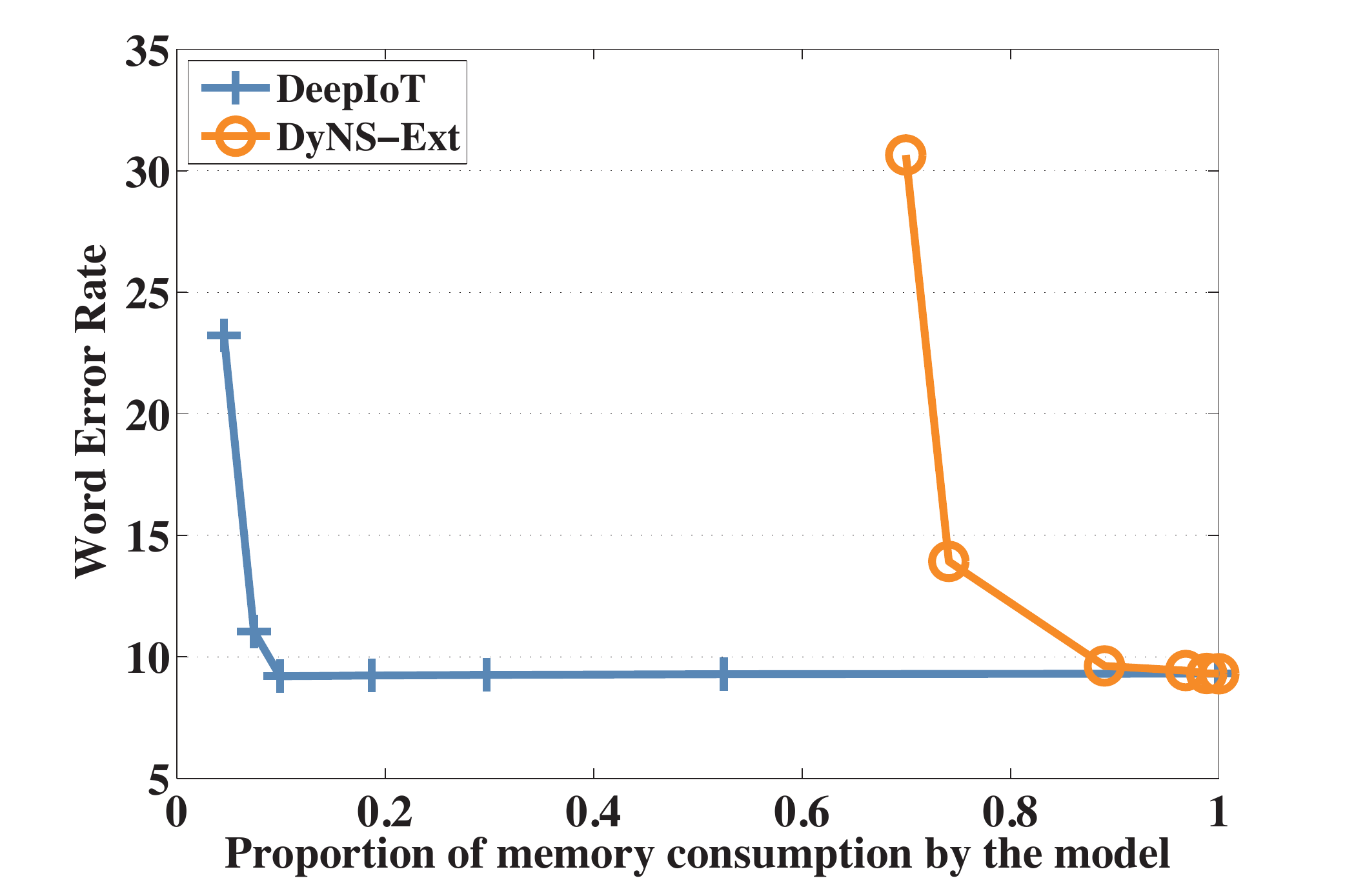}
  \caption{The tradeoff between word error rate and memory consumption by models. }
  \label{fig:lstm_acc_mem}
\end{subfigure}%
\begin{subfigure}{.32\linewidth}
  \centering
  \includegraphics[width=1.\linewidth]{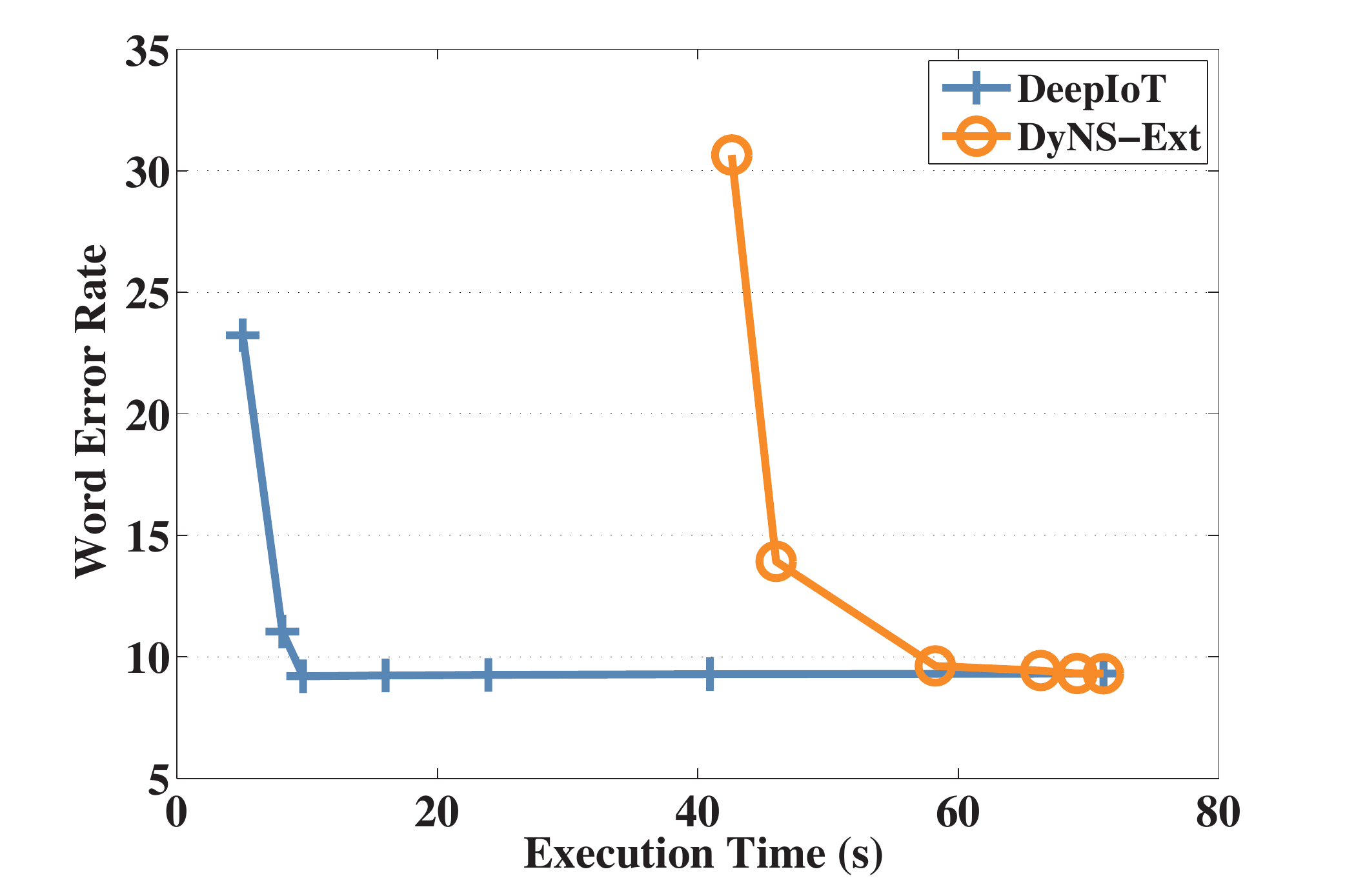}
  \caption{The tradeoff between word error rate and execution time.}
  \label{fig:lstm_acc_time}
\end{subfigure}
\begin{subfigure}{.32\linewidth}
  \centering
  \includegraphics[width=1.\linewidth]{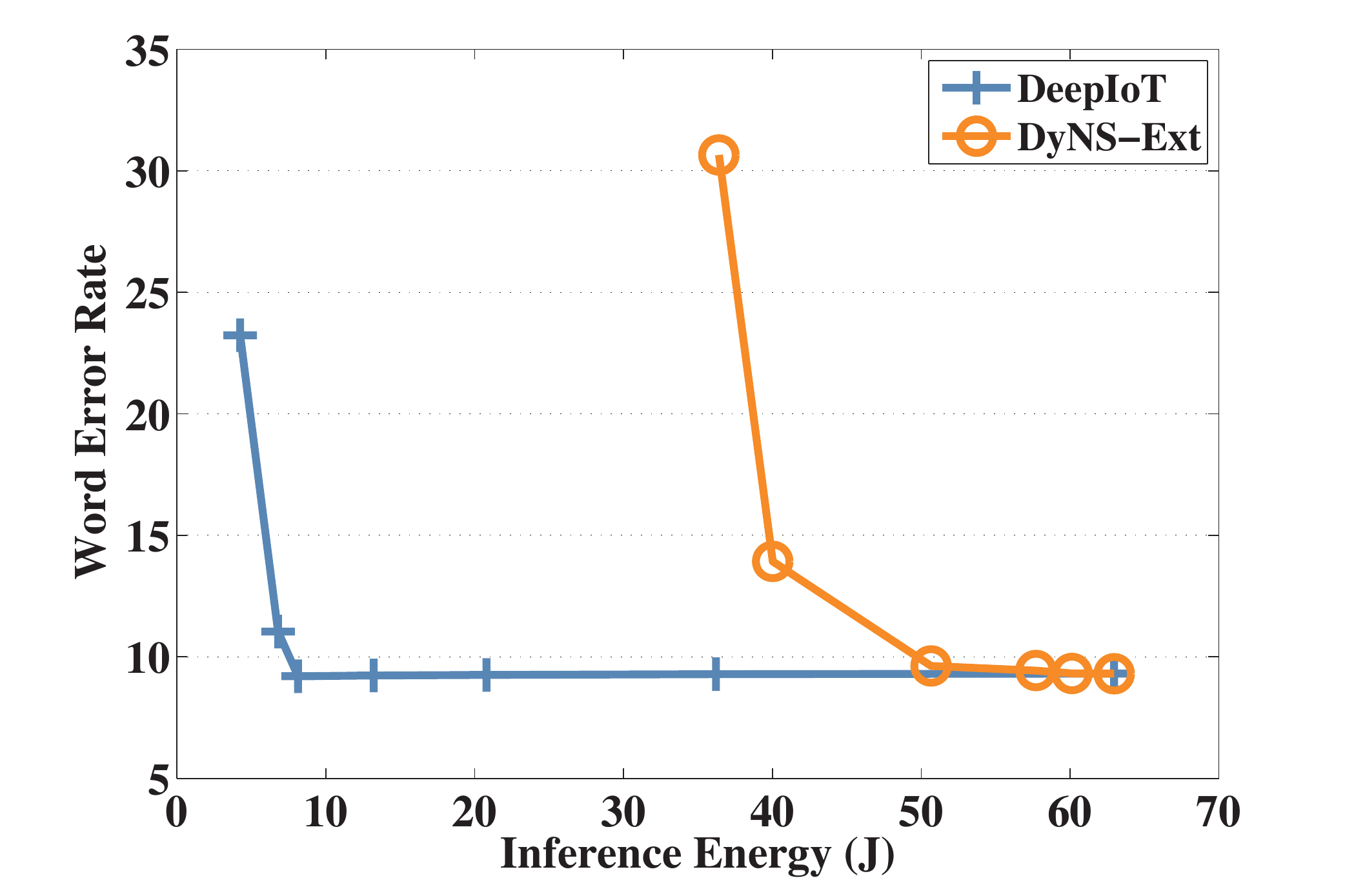}
  \caption{The tradeoff between word error rate and energy consumption.}
  \label{fig:lstm_acc_engy}
\end{subfigure}
\vspace{-0.3cm}
\caption{System performance tradeoff for deep bidirectional LSTM on LibriSpeech ASR corpus}
\label{fig:vggnet}
\vspace{-0.25cm}
\end{figure*}

Fig.~\ref{fig:vggnet_acc_time} shows the tradeoff between testing accuracy and execution time for different models. DeepIoT still achieves the best tradeoff. DeepIoT takes $82.2$ms for a prediction, which reduces $94.5\%$ execution time without the loss of accuracy. Different from the experiment with LeNet-5 on MNIST, DyNS uses less execution time compared with SparseSep in this experiment. There are two reasons for this. One is that VGGNet use smaller convolutional kernel compared with LeNet-5. Therefore factorizing 2d kernel into two 1d kernel helps less on reducing computation time. The other point is that SparseSep fails to compress the original network into a small size while keeping the original performance. As we mentioned before, it is because SparseSep avoids the fine-tuning.

Fig.~\ref{fig:vggnet_acc_engy} shows the tradeoff between testing accuracy and energy consumption for different models. DeepIoT reduces energy consumption by $95.7\%$ compared with the original VGGNet without loss of accuracy. It greatly helps us to develop a long-standing application with deep neural network in energy-constrained embedded devices.

\subsubsection{Speech recognition with deep Bidirectional LSTM}
The third human interaction modality is speech.
The sensing system can take the voices of users from the microphone and automatically convert what users said into text.
The previous two experiments mainly focus on the network structure with convolutional layers and fully-connected layers. We see how DeepIoT and the baseline algorithms work on the recurrent neural network in this section.

In this experiment, we use LibriSpeech ASR corpus~\cite{panayotov2015librispeech} as our training and testing dataset. The LibriSpeech ASR corpus is a large-scale corpus of read English speech. It consists of 460-hour training data and 2-hour testing data.  

We choose deep bidirectional LSTM as the original model~\cite{graves2014towards} in this experiment. It takes mel frequency cepstral coefficient (MFCC) features of voices as inputs, and uses two 5-layer long short-term memory (LSTM) in both forward and backward direction. The output of two LSTM are jointly used to predict the spoken text. The detailed network structure is shown in the first column of Table~\ref{tab:LSTM}, where ``LSTMf" denotes the LSTM in forward direction and ``LSTMb" denotes the LSTM in backward direction.

Two baseline algorithms are not applicable to the recurrent neural network, so we compared DeepIoT only with SyNS-Ext in this experiment. The word error rate (WER), defined as the edit distance between the true word sequence and the most probable word sequence predicted by the neural network, is used as the evaluation metric for this experiment.

We show the detailed statistics of best compressed model that keeps the original WER in Table~\ref{tab:LSTM}. DeepIoT achieves a significantly better compression rate compared with DyNS-Ext, and the model generated by DeepIoT even has a little improvement on WER. However, compared with the previous two examples on convolutional neural network,  DeepIoT fails to compress the model to less than $5\%$ of the original parameters in the recurrent neural network case (still a 20-fold reduction though). The main reason is that compressing recurrent networks needs to prune both the output dimension and the hidden dimension. It has been shown that dropping hidden dimension can harm the network performance~\cite{zaremba2014recurrent}. However DeepIoT is still successful in compressing network to less than $10\%$ of parameters.

Fig.~\ref{fig:lstm_acc_mem} shows the tradeoff between word error rate and memory consumption by compressed models. DeepIoT achieves around $\times 7$ better tradeoff compared with magnitude-based method, DyNS-Ext. This means compressing recurrent neural networks requires more information about parameter redundancies within and among each layer. Compression using only local information, such as magnitude information, will cause degradation in the final performance.

\begin{table*}[!ht]
\footnotesize
\begin{center}
\small
\caption {Heterogeneous human activity recognition}
\vspace{-0.3cm}
\label{tab:HHAR}
\begin{tabular}{ |c| c | c | c | c | c | c | c | c | c | c | c | c | c | c | c |} 
 \hline
 \multicolumn{2}{ |c| }{Layer} & \multicolumn{2}{ c| }{Hidden Unit} & \multicolumn{2}{ c| }{Params} & \multicolumn{4}{ c| }{DeepIoT (Hidden Units/ Params)} & \multicolumn{2}{ c| }{DyNS-Ext} & \multicolumn{2}{ c| }{DyNS} & \multicolumn{2}{ c| }{SparseSep} \\ 
  \hline
   \hline
  conv1a & conv1b $(2\times 9)$  & $\phantom{a}$64$\phantom{a}$ & 64 & 1.1K & 1.1K & 20 & 19   & $\phantom{1}$31.25\%$\phantom{1}$ & 29.69\% & 92\% & 95.7\% & 50.3\% & 60.0\% & 100\% & 100\%\\ 
  \hline
  conv2a & conv2b $(1\times 3)$  & $\phantom{a}$64$\phantom{a}$ & 64 & 12.3K & 12.3K & 20 & 14 & 9.76\% & 6.49\% & 70.1\% & 77.7\% & 25.3\% & 40.5\% & 114\% & 114\% \\ 
  \hline
  conv3a & conv3b $(1\times 3)$  & $\phantom{a}$64$\phantom{a}$ & 64 & 12.3K & 12.3K & 23 & 23 & 11.23\%& 7.86\% & 69.9\% & 66.2\% & 32.1\% & 35.4\% & 114\% & 114\% \\ 
  \hline
   \multicolumn{2}{ |c| }{conv4 $(2\times 8)$} & \multicolumn{2}{ c| }{64} &  \multicolumn{2}{ c| }{65.5K} & \multicolumn{2}{ c| }{10} & \multicolumn{2}{ c| }{5.61\%}& \multicolumn{2}{ c| }{40.3\%} & \multicolumn{2}{ c| }{20.4\%} & \multicolumn{2}{ c| }{53.7\%}\\ 
  \hline
  \multicolumn{2}{ |c| }{conv5 $(1\times 6)$} & \multicolumn{2}{ c| }{64} & \multicolumn{2}{ c| }{24.6K} & \multicolumn{2}{ c| }{12} & \multicolumn{2}{ c| }{2.93\%}& \multicolumn{2}{ c| }{27.2\%} & \multicolumn{2}{ c| }{18.3\%} & \multicolumn{2}{ c| }{100\%}\\ 
  \hline
  \multicolumn{2}{ |c| }{conv6 $(1\times 4)$} & \multicolumn{2}{ c| }{64} & \multicolumn{2}{ c| }{16.4K} & \multicolumn{2}{ c| }{17} & \multicolumn{2}{ c| }{4.98\%}& \multicolumn{2}{ c| }{24.6\%} & \multicolumn{2}{ c| }{12.0\%} & \multicolumn{2}{ c| }{100\%}\\ 
  \hline
  \multicolumn{2}{ |c| }{gru1 } & \multicolumn{2}{ c| }{120} & \multicolumn{2}{ c| }{227.5K} & \multicolumn{2}{ c| }{27} & \multicolumn{2}{ c| }{5.8\%}& \multicolumn{2}{ c| }{1.2\%} & \multicolumn{2}{ c| }{100\%} & \multicolumn{2}{ c| }{100\%}\\ 
  \hline
   \multicolumn{2}{ |c| }{gru2 } & \multicolumn{2}{ c| }{120}& \multicolumn{2}{ c| }{86.4K} & \multicolumn{2}{ c| }{31} & \multicolumn{2}{ c| }{6.24\%}& \multicolumn{2}{ c| }{3.6\%} & \multicolumn{2}{ c| }{100\%} & \multicolumn{2}{ c| }{100\%}\\ 
  \hline
  \multicolumn{2}{ |c| }{fc1 } & \multicolumn{2}{ c| }{6}& \multicolumn{2}{ c| }{0.7K} & \multicolumn{2}{ c| }{6} & \multicolumn{2}{ c| }{25.83\%}& \multicolumn{2}{ c| }{98.6\%} & \multicolumn{2}{ c| }{99\%} & \multicolumn{2}{ c| }{70\%}\\ 
  \hline
   \multicolumn{2}{ |c| }{total } & \multicolumn{2}{ c| }{\diagbox[dir=SW,width=2.cm, height=0.31cm]{}{}}& \multicolumn{2}{ c| }{472.5K} & \multicolumn{2}{ c| }{\diagbox[dir=SW,width=1.3cm, height=0.31cm]{}{}} & \multicolumn{2}{ c| }{6.16\%}& \multicolumn{2}{ c| }{17.1\%} & \multicolumn{2}{ c| }{74.5\%} & \multicolumn{2}{ c| }{95.3\%}\\ 
  \hline
  \hline
   \multicolumn{2}{ |c| }{Test Accuracy} & \multicolumn{4}{ c| }{94.6\%} & \multicolumn{4}{ c| }{94.7\%} & \multicolumn{2}{ c| }{94.6\%} & \multicolumn{2}{ c| }{94.6\%} & \multicolumn{2}{ c| }{93.7\%}\\
    \hline
\end{tabular}
\end{center}
\end{table*}

\begin{table*}[!ht]
\small
\vspace{-0.3cm}
\begin{center}
\footnotesize
\caption {User identification with biometric motion analysis}
\vspace{-0.3cm}
\label{tab:UserID}
\begin{tabular}{ |c| c | c | c | c | c | c | c | c | c | c | c | c | c | c | c |} 
 \hline
 \multicolumn{2}{ |c| }{Layer} & \multicolumn{2}{ c| }{Hidden Unit} & \multicolumn{2}{ c| }{Params} & \multicolumn{4}{ c| }{DeepIoT (Hidden Units/ Params)} & \multicolumn{2}{ c| }{DyNS-Ext}& \multicolumn{2}{ c| }{DyNS} & \multicolumn{2}{ c| }{SparseSep} \\ 
  \hline
   \hline
  conv1a  & conv1b $(2\times 9)$  & $\phantom{a}$64$\phantom{a}$ & 64 & 1.1K & 1.1K & 7 & 1  & $\phantom{1}$10.93\%$\phantom{1}$ & 1.56\% & 64.4\% & 75.5\%& 66.8\% & 65.6\%& 100\% & 100\%\\ 
  \hline
  conv2a  & conv2b $(1\times 3)$  & $\phantom{a}$64$\phantom{a}$ & 64 & 12.3K & 12.3K & 7 & 4 & 1.2\% & 0.1\% & 32.5\% & 34.7\% & 36.6\% & 48.0\% & 114\% & 114\% \\ 
  \hline
  conv3a  & conv3b $(1\times 3)$  & $\phantom{a}$64$\phantom{a}$ & 64 & 12.3K & 12.3K & 9 & 9 & 1.54\%& 0.88\% & 31.6\% & 28.6\% & 38.4\% & 43.5\%  & 114\% & 114\% \\ 
  \hline
   \multicolumn{2}{ |c| }{conv4 $(2\times 8)$} & \multicolumn{2}{ c| }{64} &  \multicolumn{2}{ c| }{65.5K} & \multicolumn{2}{ c| }{7} & \multicolumn{2}{ c| }{1.54\%}& \multicolumn{2}{ c| }{12.1\%}& \multicolumn{2}{ c| }{29.2\%}& \multicolumn{2}{ c| }{53.7\%}\\ 
  \hline
  \multicolumn{2}{ |c| }{conv5 $(1\times 6)$} & \multicolumn{2}{ c| }{64} & \multicolumn{2}{ c| }{24.6K} & \multicolumn{2}{ c| }{5} & \multicolumn{2}{ c| }{0.85\%}& \multicolumn{2}{ c| }{21.0\%}& \multicolumn{2}{ c| }{23.3\%}& \multicolumn{2}{ c| }{100\%}\\ 
  \hline
  \multicolumn{2}{ |c| }{conv6 $(1\times 4)$} & \multicolumn{2}{ c| }{64} & \multicolumn{2}{ c| }{16.4K} & \multicolumn{2}{ c| }{7} & \multicolumn{2}{ c| }{0.85\%}& \multicolumn{2}{ c| }{18.9\%}& \multicolumn{2}{ c| }{16.0\%}& \multicolumn{2}{ c| }{100\%}\\ 
  \hline
  \multicolumn{2}{ |c| }{gru1 } & \multicolumn{2}{ c| }{120} & \multicolumn{2}{ c| }{227.5K} & \multicolumn{2}{ c| }{13} & \multicolumn{2}{ c| }{1.18\%}& \multicolumn{2}{ c| }{0.42\%}& \multicolumn{2}{ c| }{100\%}& \multicolumn{2}{ c| }{100\%}\\ 
  \hline
   \multicolumn{2}{ |c| }{gru2 } & \multicolumn{2}{ c| }{120}& \multicolumn{2}{ c| }{86.4K} & \multicolumn{2}{ c| }{9} & \multicolumn{2}{ c| }{0.69\%}& \multicolumn{2}{ c| }{1.61\%}& \multicolumn{2}{ c| }{100\%}& \multicolumn{2}{ c| }{100\%}\\ 
  \hline
  \multicolumn{2}{ |c| }{fc1 } & \multicolumn{2}{ c| }{9}& \multicolumn{2}{ c| }{1.1K} & \multicolumn{2}{ c| }{9} & \multicolumn{2}{ c| }{7.5\%}& \multicolumn{2}{ c| }{89.6\%}& \multicolumn{2}{ c| }{98\%}& \multicolumn{2}{ c| }{88\%}\\ 
  \hline
   \multicolumn{2}{ |c| }{total } & \multicolumn{2}{ c| }{\diagbox[dir=SW,width=2.cm, height=0.31cm]{}{}}& \multicolumn{2}{ c| }{472.9K} & \multicolumn{2}{ c| }{\diagbox[dir=SW,width=1.3cm, height=0.31cm]{}{}} & \multicolumn{2}{ c| }{1.13\%}& \multicolumn{2}{ c| }{7.76\%}& \multicolumn{2}{ c| }{77.0\%}& \multicolumn{2}{ c| }{95.4\%}\\ 
  \hline
  \hline
   \multicolumn{2}{ |c| }{Test Accuracy} & \multicolumn{4}{ c| }{99.6\%} & \multicolumn{4}{ c| }{99.6\%} & \multicolumn{2}{ c| }{99.6\%} & \multicolumn{2}{ c| }{99.6\%} & \multicolumn{2}{ c| }{98.8\%}\\
    \hline
    \end{tabular}
\end{center}
\end{table*}

\begin{figure*}[!ht]
\vspace{-0.3cm}
\begin{subfigure}{.32\linewidth}
  \centering
  \includegraphics[width=1.\linewidth]{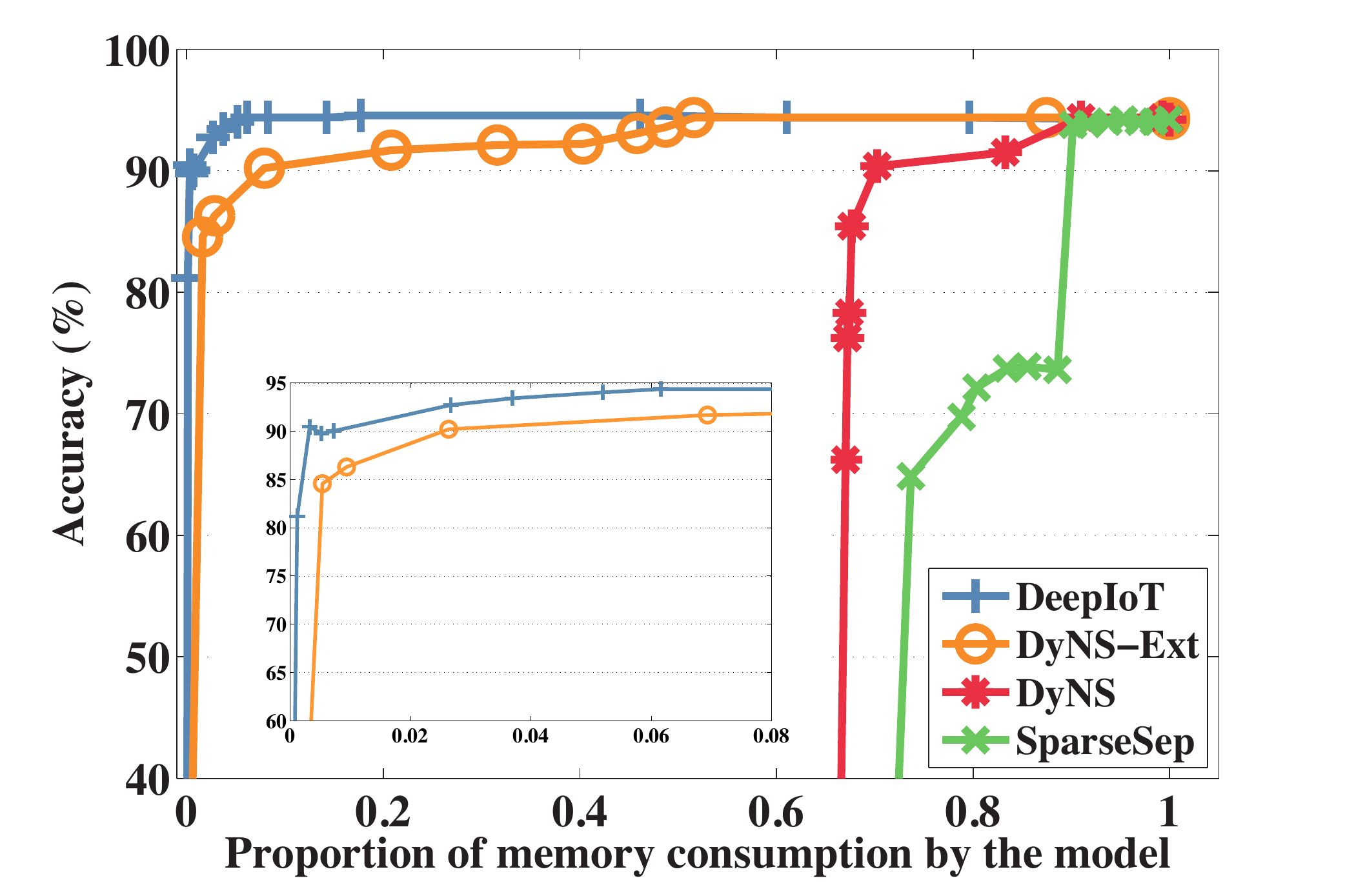}
  \caption{The tradeoff between testing accuracy and memory consumption by models. }
  \label{fig:HHAR-mem}
\end{subfigure}%
\begin{subfigure}{.32\linewidth}
  \centering
  \includegraphics[width=1.\linewidth]{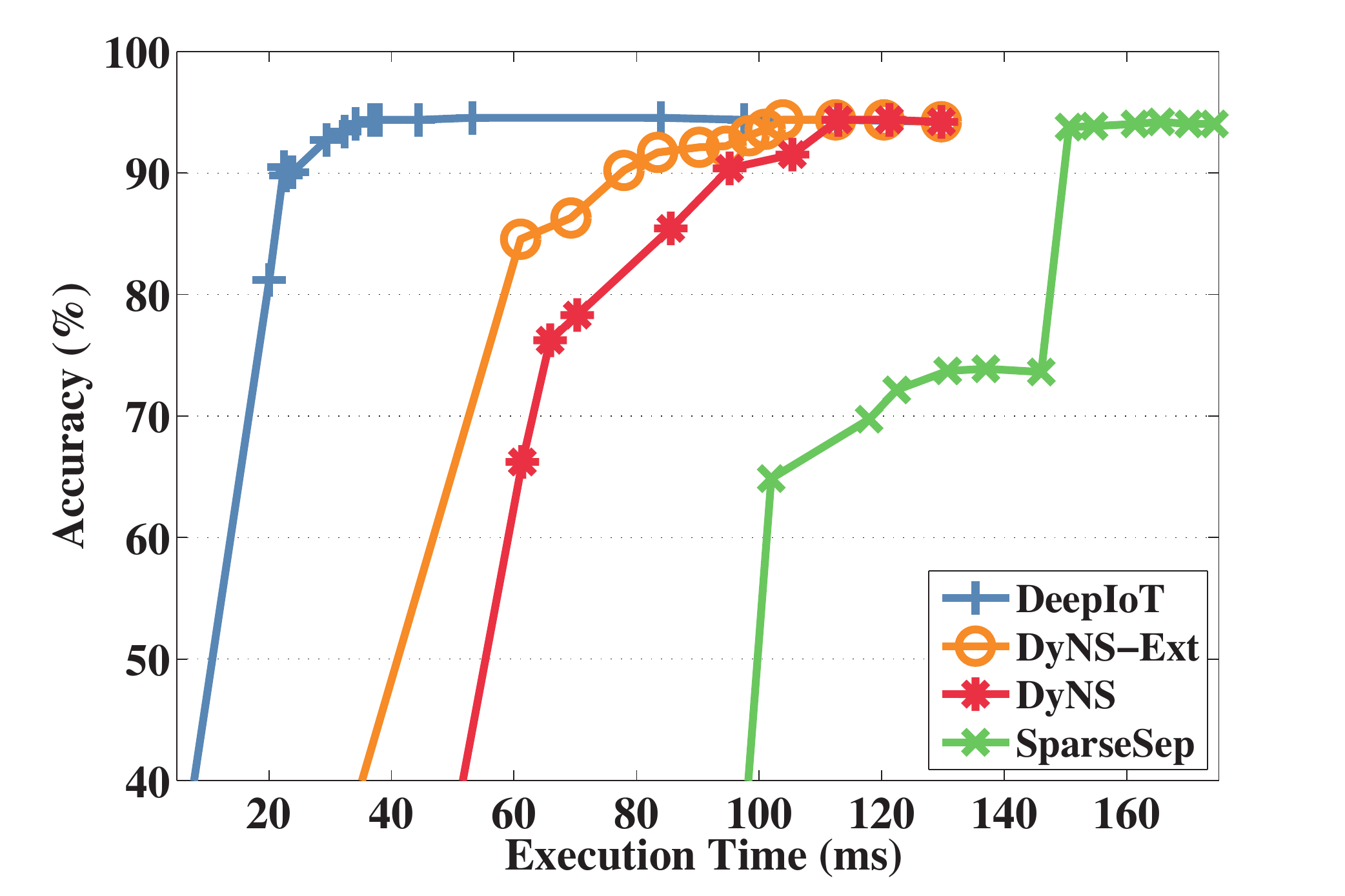}
  \caption{The tradeoff between testing accuracy and execution time.}
  \label{fig:HHAR-time}
\end{subfigure}
\begin{subfigure}{.32\linewidth}
  \centering
  \includegraphics[width=1.\linewidth]{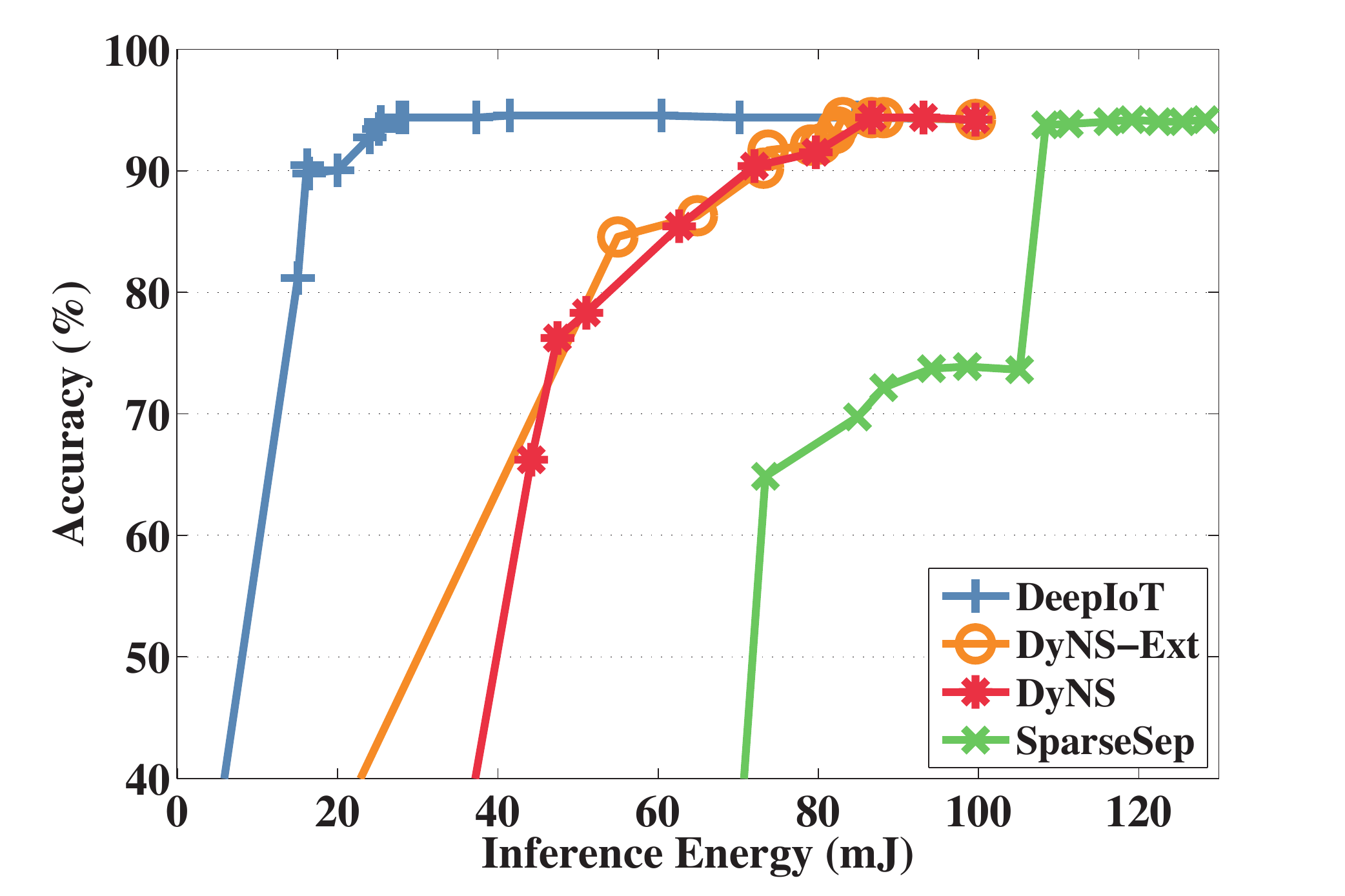}
  \caption{The tradeoff between testing accuracy and energy consumption.}
  \label{fig:HHAR-engy}
\end{subfigure}
\vspace{-0.3cm}
\caption{System performance tradeoff for heterogeneous human activity recognition}
\label{fig:HHAR}
\vspace{-0.3cm}
\end{figure*}

\begin{figure*}[!ht]
\begin{subfigure}{.32\linewidth}
  \centering
  \includegraphics[width=1.\linewidth]{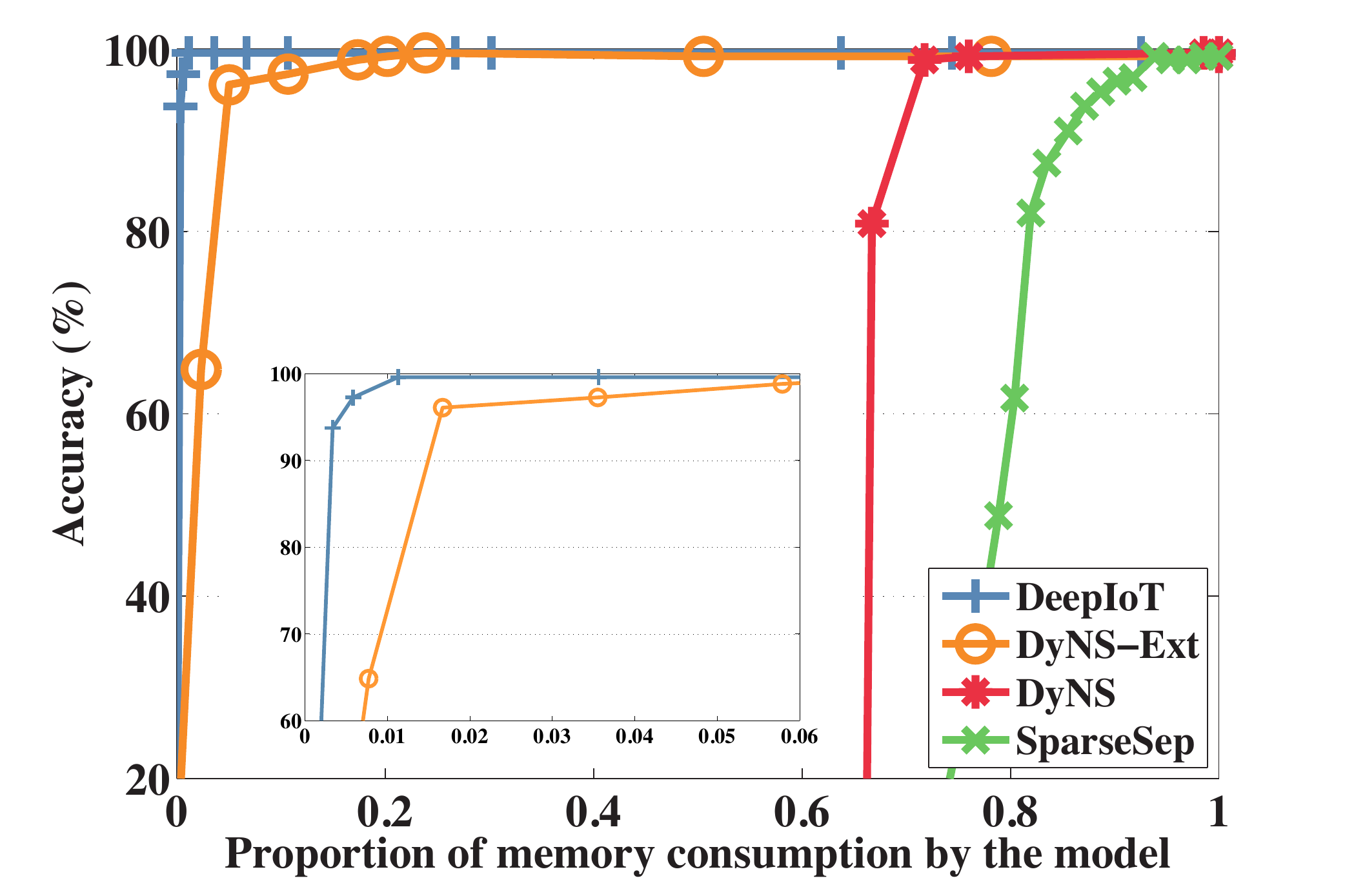}
  \caption{The tradeoff between testing accuracy and memory consumption by models. }
  \label{fig:UserID-mem}
\end{subfigure}%
\begin{subfigure}{.32\linewidth}
  \centering
  \includegraphics[width=1.\linewidth]{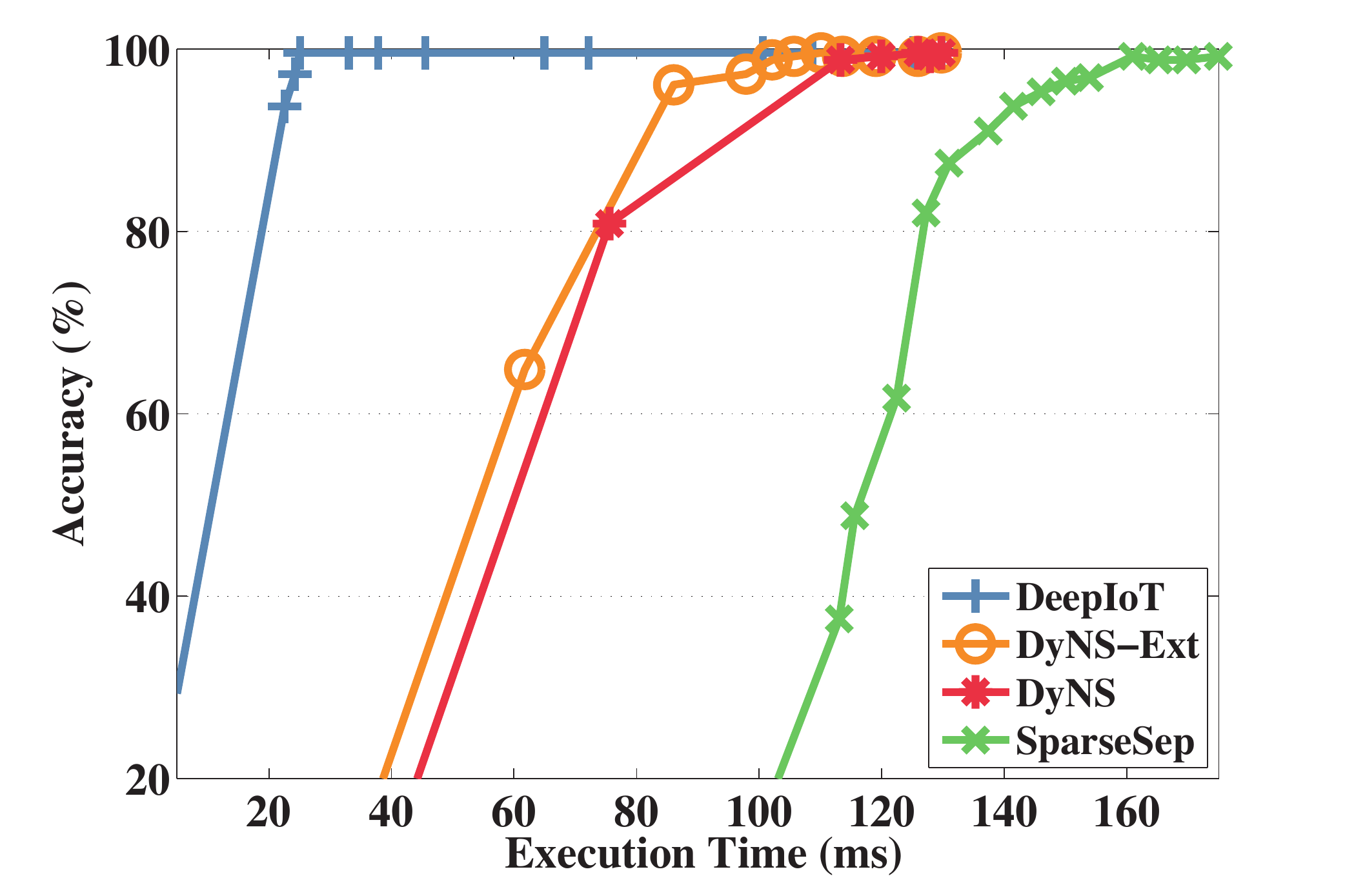}
  \caption{The tradeoff between testing accuracy and execution time.}
  \label{fig:UserID-time}
\end{subfigure}
\begin{subfigure}{.32\linewidth}
  \centering
  \includegraphics[width=1.\linewidth]{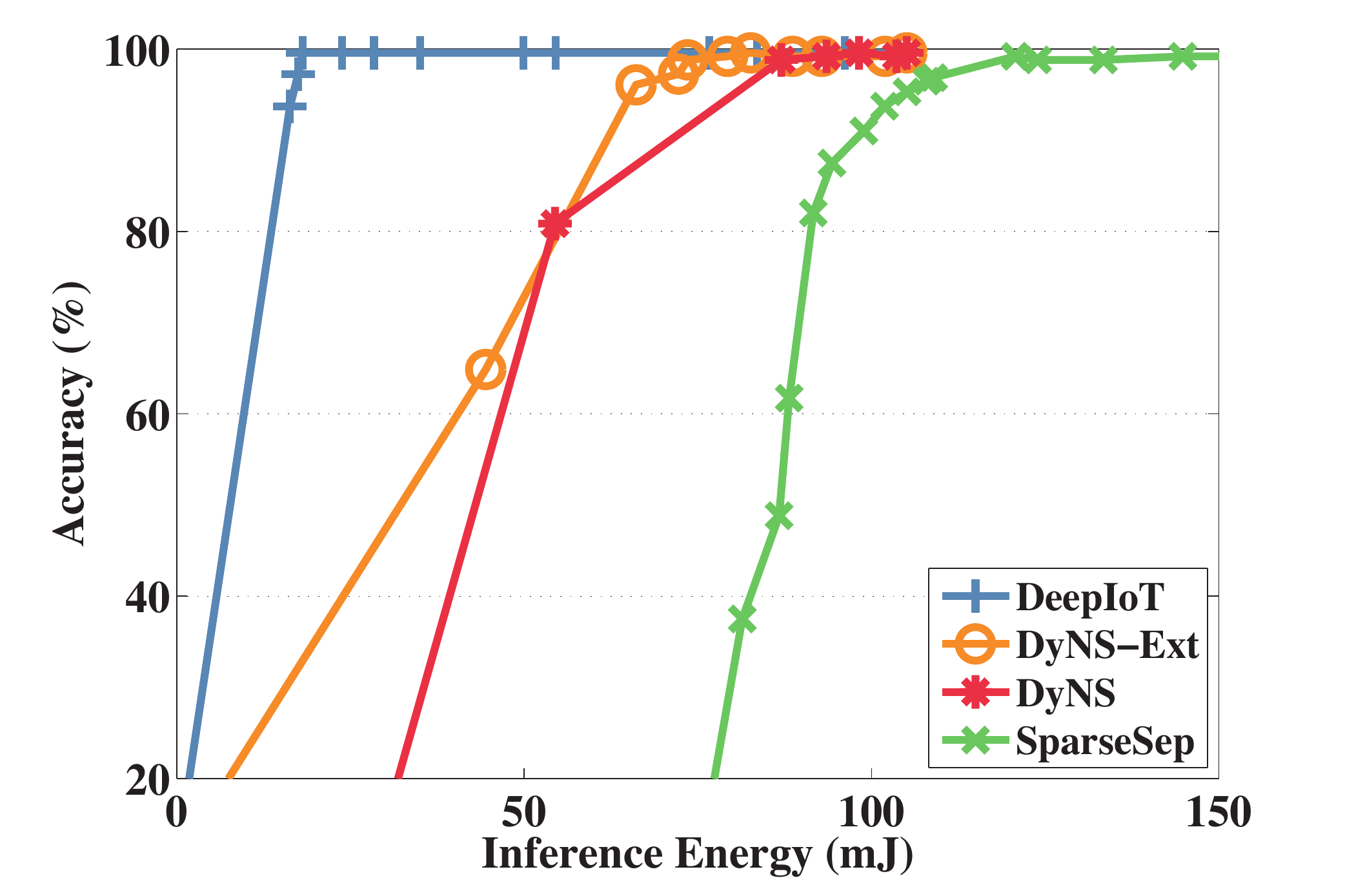}
  \caption{The tradeoff between testing accuracy and energy consumption.}
  \label{fig:UserID-engy}
\end{subfigure}
\vspace{-0.3cm}
\caption{System performance tradeoff for user identification with biometric motion analysis}
\vspace{-0.3cm}
\label{fig:UserID}
\vspace{-0.25cm}
\end{figure*}

Fig.~\ref{fig:lstm_acc_time} shows the tradeoff between word error rate and execution time. DeepIoT reduces execution time by $86.4\%$ without degradation on WER compared with the original network. With the evaluation on Intel Edision, the original network requires $71.15$ seconds in average to recognize one human speak voice example with the average length of $7.43$ seconds. The compressed structure generated by DeepIoT reduces the average execution time to $9.68$ seconds without performance loss, 
which improves responsiveness of human voice recognition.

Fig.~\ref{fig:lstm_acc_engy} shows the tradeoff between word error rate and energy consumption. DeepIoT reduces energy by $87\%$ compared with the original network. It performs better than DyNS-Ext by a large margin.

\vspace{-0.25cm}
\subsection{Supporting Human-Centric Context Sensing}

In addition to experiments about supporting basic human-centric interaction modalities, we evaluate DeepIoT on several human-centric context sensing applications. We compress the state-of-the-art deep learning model, DeepSense,~\cite{yao2017deepsense} for these problems and evaluate the accuracy and other system performance for the compressed networks. DeepSense contains all commonly used modules, including convolutional, recurrent, and fully-connected layers, which is also a good example to test the performance of compression algorithms on the combination of different types of neural network modules.

Two human-centric context sensing tasks we consider are heterogeneous human activity recognition (HHAR) and user identification with biometric motion analysis (UserID). The HHAR task recognizes human activities with motion sensors, accelerometer and gyroscope. ``Heterogeneous" means that the task is focus on the generalization ability with human who has not appeared in the training set. The UserID task identifies users during a person's daily activities such as walking, running, siting, and standing.

In this evaluation section, we use the dataset collected by Allan et al.~\cite{stisen2015smart}. This dataset contains readings from two motion sensors (accelerometer and gyroscope). Readings were recorded when users execute activities scripted in no specific order, while carrying smartwatches and smartphones. The dataset contains 9 users, 6 activities (biking, sitting, standing, walking, climbStairup, and climbStair-down), and 6 types of mobile devices. For both tasks, accelerometer and gyroscope measurements are model inputs. However, for HHAR tasks, activities are used as labels, and for UserID tasks, users' unique IDs are used as labels.

The original network structure of DeepSense is shown in the first two columns of Table~\ref{tab:HHAR} and~\ref{tab:UserID}. Both tasks use a unified neural network structure as introduced in~\cite{yao2017deepsense}. The structure contains both convolutional and recurrent layers. 
Since SparseSep and DyNS are not directly applicable to recurrent layers, we keep the recurrent layers unchanged while using them.
In addition, we also compare DeepIoT with DyNS-Ext in this experiment.

Table~\ref{tab:HHAR} and~\ref{tab:UserID} illustrate the statistics of final pruned network generated by four algorithms that have no or acceptable degradation on testing accuracy. DeepIoT is the best-performing algorithm considering the remaining number of network parameters. This is mainly due to the design of compressor network and compressor-critic framework that jointly reduce the redundancies among parameters while maintaining a global view across different layers. DyNS and SparseSep are two algorithms that can be only applied to the fully-connected and convolutional layers in the original structure. Therefore there exists a lower bound of the left proportion of parameters, \ie the number of parameters in recurrent layers. This lower bound is around $66 \%$.

The detailed tradeoffs between testing accuracy and memory consumption by the models are illustrated in Fig.~\ref{fig:HHAR-mem} and~\ref{fig:UserID-mem}. DeepIoT still achieves the best tradeoff for sensing applications. Other than the compressor neural network providing global parameter redundancies, directly pruning hidden elements in each layer also enables DeepIoT to obtain more concise representations in matrix form, which results in less memory consumption. 

The tradeoffs between system execution time and testing accuracy are shown in Fig.~\ref{fig:HHAR-time} and~\ref{fig:UserID-time}. DeepIoT uses the least execution time when achieving the same testing accuracy compared with three baselines. It takes $36.7$ms and $25.1$ms for a single prediction, which reduces execution time by around $80.8\%$ and $71.4\%$ in UserID and HHAR, respectively, without loss of accuracy. DyNS and DyNS-Ext achieve better performance on time compared with SparseSep, which is different from the previous evaluations on LeNet-5. It is the structure of the original neural network that causes this difference. As shown in Table~\ref{tab:HHAR} and~\ref{tab:UserID}, the original network uses 1-d filters in its structure. The matrix factorization based kernel compressing method used in SparseSep cannot help to reduce or even increase the parameter redundancies and the number of operations involved. Therefore, there are constraints on the network structure when applying matrix factorization based compression algorithm. In addition, SparseSep cannot be applied to the recurrent layers in the network, which consumes a large proportion of operations during running the neural network. 

The tradeoffs between energy consumption and testing accuracy are shown in Fig.~\ref{fig:HHAR-engy} and~\ref{fig:UserID-engy}. DeepIoT is the best-performing algorithm for energy consumption. It reduces energy by around $83.3\%$ and $72.2\%$ in UserID and HHAR without loss of accuracy.  Due to the aforementioned problem of SparseSep on 1-d filter, redundant factorization causes more execution time and energy consumption in the experiment.

\vspace{-0.2cm}
\section{Discussion}~\label{sec:discussion}
This paper tries to apply state-of-the-art neural network models on resource-constrained embedded and mobile devices by simplifying network structure without hurting accuracy. Our solution, DeepIoT, generates a simplified network structure by deciding which elements to drop in each layer. This whole process requires fine-tuning (Line 18 to Line 21 in Algorithm~\ref{alg:comp_pred}). However, we argue that the fine-tuning step should not be the obstacle in applying DeepIoT. First, all the compressing and fine-tuning steps are conducted on the workstation instead of embedded and mobile devices. We can easily apply the DeepIoT algorithm to compress and fine-tune the neural network ahead of time and then deploy the compressed model into embedded and mobile devices without any run-time processing. Second, the original training data must be easily accessible. Developers who want to apply neural networks to solve their own sensing problems will typically have access to their own datasets to fine-tune the model. For others, lots of large-scale datasets are available online, such as vision~\cite{deng2009imagenet} and audio~\cite{jort2017audioset} data. 
Hence, for many categories of applications, fine-tuning is feasible. 

DeepIoT mainly focuses on structure pruning or weight pruning, which is independent from other model compression methods such as weight quantization~\cite{gong2014compressing,courbariaux2016binarized,gupta2015deep, courbariaux2014training}. Although weight quantization can compress the network complexity by using limited numerical precision or clustering parameters, the compression ratio of quantization is usually less than the structure pruning methods. Weight pruning and quantization are two non-conflicting methods. We can apply weight quantization after the structure pruning step or after any other compression algorithm. This is out of the scope of this paper.  Also we do not exploit heterogeneous local device processors, such as DSPs~\cite{lane2016deepx}, to speed up the running time during all the experiments, because this paper focuses on structure compression methods for deep neural networks instead of hardware speed-up.

\vspace{-0.2cm}
\section{Conclusion}~\label{sec:conclusion}
In this paper, we described DeepIoT, a compression algorithm that learns a more succinct network structure for sensing applications. DeepIoT integrates the original network with dropout learning and generates stochastic hidden elements in each layer. We also described a novel compressor neural network to learn the parameter redundancies and generate dropout probabilities for original network layers. The compressor neural network is optimized jointly with the original neural network through the compressor-critic framework. DeepIoT outperforms other baseline compression algorithms by a significant margin in all experiments. The compressed structure greatly reduces the resource consumption on sensing system without hurting the performance, and makes a lot of the state-of-the-art deep neural networks deployable on resource-constrained embedded and mobile devices.

\vspace{-0.2cm}
\section{Acknowledgements}
We sincerely thank Nicholas D. Lane for shepherding the final version of this paper, and the anonymous reviewers for their invaluable comments. Research reported in this paper was sponsored in part by NSF under grants CNS 16-18627 and CNS 13-20209 and in part by the Army Research Laboratory under Cooperative Agreement W911NF-09-2-0053. The views and conclusions contained in this document are those of the authors and should not be interpreted as representing the official policies, either expressed or implied, of the Army Research Laboratory, NSF, or the U.S. Government. The U.S. Government is authorized to reproduce and distribute reprints for Government purposes notwithstanding any copyright notation here on.

}

\newpage
\bibliographystyle{abbrv}
\bibliography{reference}

\begin{thebibliography}{10}

\bibitem{Edison}
Intel edison compute module.
\newblock
  \url{http://www.intel.com/content/dam/support/us/en/documents/edison/sb/edison-module_HG_331189.pdf}.

\bibitem{ubilinux}
Loading debian (ubilinux) on the edison.
\newblock
  \url{https://learn.sparkfun.com/tutorials/loading-debian-ubilinux-on-the-edison}.

\bibitem{alsheikh2014machine}
M.~A. Alsheikh, S.~Lin, D.~Niyato, and H.-P. Tan.
\newblock Machine learning in wireless sensor networks: Algorithms, strategies,
  and applications.
\newblock {\em IEEE Communications Surveys \& Tutorials}, 16(4):1996--2018,
  2014.

\bibitem{lane2016sparsifying}
S.~Bhattacharya and N.~D. Lane.
\newblock Sparsification and separation of deep learning layers for constrained
  resource inference on wearables.
\newblock In {\em Proceedings of the 14th ACM Conference on Embedded Network
  Sensor Systems CD-ROM}, pages 176--189. ACM, 2016.

\bibitem{brouwers2014incremental}
N.~Brouwers, M.~Zuniga, and K.~Langendoen.
\newblock Incremental wi-fi scanning for energy-efficient localization.
\newblock In {\em Pervasive Computing and Communications (PerCom), 2014 IEEE
  International Conference on}, pages 156--162. IEEE, 2014.

\bibitem{capra2003carisma}
L.~Capra, W.~Emmerich, and C.~Mascolo.
\newblock Carisma: Context-aware reflective middleware system for mobile
  applications.
\newblock {\em IEEE Transactions on software engineering}, 29(10):929--945,
  2003.

\bibitem{cho2011inferring}
E.~Cho, K.~Wong, O.~Gnawali, M.~Wicke, and L.~Guibas.
\newblock Inferring mobile trajectories using a network of binary proximity
  sensors.
\newblock In {\em Sensor, Mesh and Ad Hoc Communications and Networks (SECON),
  2011 8th Annual IEEE Communications Society Conference on}, pages 188--196.
  IEEE, 2011.

\bibitem{clark2017vinet}
R.~Clark, S.~Wang, H.~Wen, A.~Markham, and N.~Trigoni.
\newblock Vinet: Visual inertial odometry as a sequence to sequence learning
  problem.
\newblock In {\em AAAI Conference on Artificial Intelligence (AAAI)}, 2017.

\bibitem{courbariaux2014training}
M.~Courbariaux, Y.~Bengio, and J.-P. David.
\newblock Training deep neural networks with low precision multiplications.
\newblock {\em arXiv preprint arXiv:1412.7024}, 2014.

\bibitem{courbariaux2016binarized}
M.~Courbariaux, I.~Hubara, D.~Soudry, R.~El-Yaniv, and Y.~Bengio.
\newblock Binarized neural networks: Training deep neural networks with weights
  and activations constrained to+ 1 or-1.
\newblock {\em arXiv preprint arXiv:1602.02830}, 2016.

\bibitem{deng2009imagenet}
J.~Deng, W.~Dong, R.~Socher, L.-J. Li, K.~Li, and L.~Fei-Fei.
\newblock Imagenet: A large-scale hierarchical image database.
\newblock In {\em Computer Vision and Pattern Recognition, 2009. CVPR 2009.
  IEEE Conference on}, pages 248--255. IEEE, 2009.

\bibitem{denton2014exploiting}
E.~L. Denton, W.~Zaremba, J.~Bruna, Y.~LeCun, and R.~Fergus.
\newblock Exploiting linear structure within convolutional networks for
  efficient evaluation.
\newblock In {\em Advances in Neural Information Processing Systems}, pages
  1269--1277, 2014.

\bibitem{ferrari2012low}
F.~Ferrari, M.~Zimmerling, L.~Mottola, and L.~Thiele.
\newblock Low-power wireless bus.
\newblock In {\em Proceedings of the 10th ACM Conference on Embedded Network
  Sensor Systems}, pages 1--14. ACM, 2012.

\bibitem{gal2015bayesian}
Y.~Gal and Z.~Ghahramani.
\newblock Bayesian convolutional neural networks with bernoulli approximate
  variational inference.
\newblock {\em arXiv preprint arXiv:1506.02158}, 2015.

\bibitem{gal2015theoretically}
Y.~Gal and Z.~Ghahramani.
\newblock A theoretically grounded application of dropout in recurrent neural
  networks.
\newblock 2016.

\bibitem{jort2017audioset}
J.~F. Gemmeke, D.~P.~W. Ellis, D.~Freedman, A.~Jansen, W.~Lawrence, R.~C.
  Moore, M.~Plakal, and M.~Ritter.
\newblock Audio set: An ontology and human-labeled dataset for audio events.
\newblock In {\em Proc. IEEE ICASSP 2017}, New Orleans, LA, 2017.

\bibitem{glynn1990likelihood}
P.~W. Glynn.
\newblock Likelihood ratio gradient estimation for stochastic systems.
\newblock {\em Communications of the ACM}, 33(10):75--84, 1990.

\bibitem{gong2014compressing}
Y.~Gong, L.~Liu, M.~Yang, and L.~Bourdev.
\newblock Compressing deep convolutional networks using vector quantization.
\newblock {\em arXiv preprint arXiv:1412.6115}, 2014.

\bibitem{goumas2008understanding}
G.~Goumas, K.~Kourtis, N.~Anastopoulos, V.~Karakasis, and N.~Koziris.
\newblock Understanding the performance of sparse matrix-vector multiplication.
\newblock In {\em Parallel, Distributed and Network-Based Processing, 2008. PDP
  2008. 16th Euromicro Conference on}, pages 283--292. IEEE, 2008.

\bibitem{graves2014towards}
A.~Graves and N.~Jaitly.
\newblock Towards end-to-end speech recognition with recurrent neural networks.
\newblock In {\em ICML}, volume~14, pages 1764--1772, 2014.

\bibitem{gu2015muprop}
S.~Gu, S.~Levine, I.~Sutskever, and A.~Mnih.
\newblock Muprop: Unbiased backpropagation for stochastic neural networks.
\newblock {\em arXiv preprint arXiv:1511.05176}, 2015.

\bibitem{guo2016dynamic}
Y.~Guo, A.~Yao, and Y.~Chen.
\newblock Dynamic network surgery for efficient dnns.
\newblock In {\em Advances In Neural Information Processing Systems}, pages
  1379--1387, 2016.

\bibitem{gupta2015deep}
S.~Gupta, A.~Agrawal, K.~Gopalakrishnan, and P.~Narayanan.
\newblock Deep learning with limited numerical precision.
\newblock In {\em ICML}, pages 1737--1746, 2015.

\bibitem{han2015deep}
S.~Han, H.~Mao, and W.~J. Dally.
\newblock Deep compression: Compressing deep neural network with pruning,
  trained quantization and huffman coding.
\newblock {\em CoRR, abs/1510.00149}, 2, 2015.

\bibitem{hester2016amulet}
J.~Hester, T.~Peters, T.~Yun, R.~Peterson, J.~Skinner, B.~Golla, K.~Storer,
  S.~Hearndon, K.~Freeman, S.~Lord, et~al.
\newblock Amulet: An energy-efficient, multi-application wearable platform.
\newblock In {\em Proceedings of the 14th ACM Conference on Embedded Network
  Sensor Systems CD-ROM}, pages 216--229. ACM, 2016.

\bibitem{hinton2015distilling}
G.~Hinton, O.~Vinyals, and J.~Dean.
\newblock Distilling the knowledge in a neural network.
\newblock {\em arXiv preprint arXiv:1503.02531}, 2015.

\bibitem{hoque2014vocal}
E.~Hoque, R.~F. Dickerson, and J.~A. Stankovic.
\newblock Vocal-diary: A voice command based ground truth collection system for
  activity recognition.
\newblock In {\em Proceedings of the Wireless Health 2014 on National
  Institutes of Health}, pages 1--6. ACM, 2014.

\bibitem{konda1999actor}
V.~R. Konda and J.~N. Tsitsiklis.
\newblock Actor-critic algorithms.
\newblock In {\em NIPS}, volume~13, pages 1008--1014, 1999.

\bibitem{kusy2007tracking}
B.~Kusy, A.~Ledeczi, and X.~Koutsoukos.
\newblock Tracking mobile nodes using rf doppler shifts.
\newblock In {\em Proceedings of the 5th international conference on Embedded
  networked sensor systems}, pages 29--42. ACM, 2007.

\bibitem{lane2016deepx}
N.~D. Lane, S.~Bhattacharya, P.~Georgiev, C.~Forlivesi, L.~Jiao, L.~Qendro, and
  F.~Kawsar.
\newblock Deepx: A software accelerator for low-power deep learning inference
  on mobile devices.
\newblock In {\em Information Processing in Sensor Networks (IPSN), 2016 15th
  ACM/IEEE International Conference on}, pages 1--12. IEEE, 2016.

\bibitem{lane2015deepear}
N.~D. Lane, P.~Georgiev, and L.~Qendro.
\newblock Deepear: robust smartphone audio sensing in unconstrained acoustic
  environments using deep learning.
\newblock In {\em Proceedings of the 2015 ACM International Joint Conference on
  Pervasive and Ubiquitous Computing}, pages 283--294. ACM, 2015.

\bibitem{mnih2014neural}
A.~Mnih and K.~Gregor.
\newblock Neural variational inference and learning in belief networks.
\newblock 2014.

\bibitem{mnih2013playing}
V.~Mnih, K.~Kavukcuoglu, D.~Silver, A.~A. Rusu, J.~Veness, M.~G. Bellemare,
  A.~Graves, M.~A. Riedmiller, A.~Fidjeland, G.~Ostrovski, S.~Petersen,
  C.~Beattie, A.~Sadik, I.~Antonoglou, H.~King, D.~Kumaran, D.~Wierstra,
  S.~Legg, and D.~Hassabis.
\newblock Human-level control through deep reinforcement learning.
\newblock {\em Nature}, 518:529--533, 2015.

\bibitem{nirjon2012musicalheart}
S.~Nirjon, R.~F. Dickerson, Q.~Li, P.~Asare, J.~A. Stankovic, D.~Hong,
  B.~Zhang, X.~Jiang, G.~Shen, and F.~Zhao.
\newblock Musicalheart: A hearty way of listening to music.
\newblock In {\em Proceedings of the 10th ACM Conference on Embedded Network
  Sensor Systems}, pages 43--56. ACM, 2012.

\bibitem{panayotov2015librispeech}
V.~Panayotov, G.~Chen, D.~Povey, and S.~Khudanpur.
\newblock Librispeech: an asr corpus based on public domain audio books.
\newblock In {\em Acoustics, Speech and Signal Processing (ICASSP), 2015 IEEE
  International Conference on}, pages 5206--5210. IEEE, 2015.

\bibitem{peters2006policy}
J.~Peters and S.~Schaal.
\newblock Policy gradient methods for robotics.
\newblock In {\em 2006 IEEE/RSJ International Conference on Intelligent Robots
  and Systems}, pages 2219--2225. IEEE, 2006.

\bibitem{radu2016towards}
V.~Radu, N.~D. Lane, S.~Bhattacharya, C.~Mascolo, M.~K. Marina, and F.~Kawsar.
\newblock Towards multimodal deep learning for activity recognition on mobile
  devices.
\newblock In {\em Proceedings of the 2016 ACM International Joint Conference on
  Pervasive and Ubiquitous Computing: Adjunct}, pages 185--188. ACM, 2016.

\bibitem{rosa2017leveraging}
S.~Rosa, X.~Lu, H.~Wen, and N.~Trigoni.
\newblock Leveraging user activities and mobile robots for semantic mapping and
  user localization.
\newblock In {\em Proceedings of the Companion of the 2017 ACM/IEEE
  International Conference on Human-Robot Interaction}, pages 267--268. ACM,
  2017.

\bibitem{rowe2010contactless}
A.~Rowe, M.~Berges, and R.~Rajkumar.
\newblock Contactless sensing of appliance state transitions through variations
  in electromagnetic fields.
\newblock In {\em Proceedings of the 2nd ACM workshop on embedded sensing
  systems for energy-efficiency in building}, pages 19--24. ACM, 2010.

\bibitem{saifullah2016snow}
A.~Saifullah, M.~Rahman, D.~Ismail, C.~Lu, R.~Chandra, and J.~Liu.
\newblock Snow: Sensor network over white spaces.
\newblock In {\em Proceedings of the International Conference on Embedded
  Networked Sensor Systems (ACM SenSys)}, 2016.

\bibitem{schuss2017competition}
M.~Schuss, C.~A. Boano, M.~Weber, and K.~Roemer.
\newblock A competition to push the dependability of low-power wireless
  protocols to the edge.
\newblock In {\em Proceedings of the 14th International Conference on Embedded
  Wireless Systems and Networks (EWSN). Uppsala, Sweden}, 2017.

\bibitem{shen2014face}
Y.~Shen, W.~Hu, M.~Yang, B.~Wei, S.~Lucey, and C.~T. Chou.
\newblock Face recognition on smartphones via optimised sparse representation
  classification.
\newblock In {\em Information Processing in Sensor Networks, IPSN-14
  Proceedings of the 13th International Symposium on}, pages 237--248. IEEE,
  2014.

\bibitem{silver2016mastering}
D.~Silver, A.~Huang, C.~J. Maddison, A.~Guez, L.~Sifre, G.~Van Den~Driessche,
  J.~Schrittwieser, I.~Antonoglou, V.~Panneershelvam, M.~Lanctot, et~al.
\newblock Mastering the game of go with deep neural networks and tree search.
\newblock {\em Nature}, 529(7587):484--489, 2016.

\bibitem{srivastava2014dropout}
N.~Srivastava, G.~E. Hinton, A.~Krizhevsky, I.~Sutskever, and R.~Salakhutdinov.
\newblock Dropout: a simple way to prevent neural networks from overfitting.
\newblock {\em Journal of Machine Learning Research}, 15(1):1929--1958, 2014.

\bibitem{stisen2015smart}
A.~Stisen, H.~Blunck, S.~Bhattacharya, T.~S. Prentow, M.~B. Kj{\ae}rgaard,
  A.~Dey, T.~Sonne, and M.~M. Jensen.
\newblock Smart devices are different: Assessing and mitigatingmobile sensing
  heterogeneities for activity recognition.
\newblock In {\em Proceedings of the 13th ACM Conference on Embedded Networked
  Sensor Systems}, pages 127--140. ACM, 2015.

\bibitem{sun2011pandaa}
Z.~Sun, A.~Purohit, K.~Chen, S.~Pan, T.~Pering, and P.~Zhang.
\newblock Pandaa: physical arrangement detection of networked devices through
  ambient-sound awareness.
\newblock In {\em Proceedings of the 13th international conference on
  Ubiquitous computing}, pages 425--434. ACM, 2011.

\bibitem{tai2015convolutional}
C.~Tai, T.~Xiao, Y.~Zhang, X.~Wang, et~al.
\newblock Convolutional neural networks with low-rank regularization.
\newblock {\em arXiv preprint arXiv:1511.06067}, 2015.

\bibitem{2016arXiv160502688short}
{Theano Development Team}.
\newblock {Theano: A {Python} framework for fast computation of mathematical
  expressions}.
\newblock {\em arXiv e-prints}, abs/1605.02688, May 2016.

\bibitem{wang2013corlayer}
S.~Wang, S.~M. Kim, Y.~Liu, G.~Tan, and T.~He.
\newblock Corlayer: A transparent link correlation layer for energy efficient
  broadcast.
\newblock In {\em Proceedings of the 19th annual international conference on
  Mobile computing \& networking}, pages 51--62. ACM, 2013.

\bibitem{wang2017deep}
S.~Wang, H.~Liu, P.~H. Gomes, and B.~Krishnamachari.
\newblock Deep reinforcement learning for dynamic multichannel access.
\newblock 2017.

\bibitem{wang2016cnnpack}
Y.~Wang, C.~Xu, S.~You, D.~Tao, and C.~Xu.
\newblock Cnnpack: packing convolutional neural networks in the frequency
  domain.
\newblock In {\em Advances in Neural Information Processing Systems}, pages
  253--261, 2016.

\bibitem{wang2017deepvo}
H.~Wen, S.~Wang, R.~Clark, and N.~Trigoni.
\newblock Deepvo: Towards end-to-end visual odometry with deep recurrent
  convolutional neural networks.
\newblock {\em International Conference on Robotics and Automation}, 2017.

\bibitem{wilson2011see}
J.~Wilson and N.~Patwari.
\newblock See-through walls: Motion tracking using variance-based radio
  tomography networks.
\newblock {\em IEEE Transactions on Mobile Computing}, 10(5):612--621, 2011.

\bibitem{yang2011detecting}
J.~Yang, S.~Sidhom, G.~Chandrasekaran, T.~Vu, H.~Liu, N.~Cecan, Y.~Chen,
  M.~Gruteser, and R.~P. Martin.
\newblock Detecting driver phone use leveraging car speakers.
\newblock In {\em Proceedings of the 17th annual international conference on
  Mobile computing and networking}, pages 97--108. ACM, 2011.

\bibitem{yao2016recursive}
S.~Yao, M.~T. Amin, L.~Su, S.~Hu, S.~Li, S.~Wang, Y.~Zhao, T.~Abdelzaher,
  L.~Kaplan, C.~Aggarwal, et~al.
\newblock Recursive ground truth estimator for social data streams.
\newblock In {\em Information Processing in Sensor Networks (IPSN), 2016 15th
  ACM/IEEE International Conference on}, pages 1--12. IEEE, 2016.

\bibitem{yao2017deepsense}
S.~Yao, S.~Hu, Y.~Zhao, A.~Zhang, and T.~Abdelzaher.
\newblock Deepsense: a unified deep learning framework for time-series mobile
  sensing data processing.
\newblock In {\em Proceedings of the 26th International Conference on World
  Wide Web}. International World Wide Web Conferences Steering Committee, 2017.

\bibitem{zaremba2014recurrent}
W.~Zaremba, I.~Sutskever, and O.~Vinyals.
\newblock Recurrent neural network regularization.
\newblock {\em arXiv preprint arXiv:1409.2329}, 2014.

\bibitem{zhang2017regions}
C.~Zhang, K.~Zhang, Q.~Yuan, H.~Peng, Y.~Zheng, T.~Hanratty, S.~Wang, and
  J.~Han.
\newblock Regions, periods, activities: Uncovering urban dynamics via
  cross-modal representation learning.
\newblock In {\em Proceedings of the 26th International Conference on World
  Wide Web}, pages 361--370. International World Wide Web Conferences Steering
  Committee, 2017.

\bibitem{zhang2009ocrdroid}
M.~Zhang, A.~Joshi, R.~Kadmawala, K.~Dantu, S.~Poduri, and G.~S. Sukhatme.
\newblock Ocrdroid: A framework to digitize text using mobile phones.
\newblock In {\em International Conference on Mobile Computing, Applications,
  and Services}, pages 273--292. Springer, 2009.

\end{thebibliography}

\end{document}